\documentclass[nohyperref]{article}
\usepackage{microtype}
\usepackage{graphicx}
\usepackage{subfigure}
\usepackage{booktabs} 

\usepackage[hidelinks]{hyperref}

\usepackage[accepted]{icml2023}
\usepackage{amsmath}
\usepackage{amssymb}
\usepackage{mathtools}
\usepackage{amsthm}

\theoremstyle{plain}

\theoremstyle{definition}

\theoremstyle{remark}

\usepackage[textsize=tiny]{todonotes}

\renewcommand{\paragraph}[1]{ \noindent \textbf{#1}}
\icmltitlerunning{Efficient Sparse Backpropagation  for Faster Training of Neural Networks}

\newcommand{\nnz}{\operatorname{nnz}}
\newcommand{\nnzcols}{\operatorname{cols}}
\newcommand{\nnzrows}{\operatorname{rows}}
\newcommand{\nnzvals}{\operatorname{vals}}
\newcommand{\nnzic}{\operatorname{ich}}
\newcommand{\nnzoc}{\operatorname{och}}
\newcommand{\nnzx}{\operatorname{x}}
\newcommand{\nnzy}{\operatorname{y}}
\newcommand{\avx}[1]{\textbf{\texttt{#1}}}

\newcommand{\ra}[1]{\renewcommand{\arraystretch}{#1}}
\newcommand{\zero}{\cdot}
\newcommand{\CC}{C\nolinebreak\hspace{-.05em}\raisebox{.4ex}{\tiny\bf +}\nolinebreak\hspace{-.10em}\raisebox{.4ex}{\tiny\bf +}}
\def\CC{{C\nolinebreak[4]\hspace{-.05em}\raisebox{.4ex}{\tiny\bf ++}}}

\usepackage{breakurl}

\begin{document}

\twocolumn[
\icmltitle{SparseProp: Efficient Sparse Backpropagation \\ for Faster Training of Neural Networks}

\icmlsetsymbol{equal}{*}

\begin{icmlauthorlist}
\icmlauthor{Mahdi Nikdan}{equal,ista}
\icmlauthor{Tommaso Pegolotti}{equal,eth}
\icmlauthor{Eugenia Iofinova}{ista}
\icmlauthor{Eldar Kurtic}{ista}
\icmlauthor{Dan Alistarh}{ista,nm}
\end{icmlauthorlist}

\icmlaffiliation{ista}{IST Austria}
\icmlaffiliation{eth}{ETH Zurich}
\icmlaffiliation{nm}{Neural Magic, Inc.}

\icmlcorrespondingauthor{Mahdi Nikdan}{mahdi.nikdan@ist.ac.at}
\icmlcorrespondingauthor{Tommaso Pegolotti}{tommaso.pegolotti@inf.ethz.ch}
\icmlcorrespondingauthor{Dan Alistarh}{dan.alistarh@ist.ac.at}
\icmlkeywords{Machine Learning, ICML}

\vskip 0.3in
]
\printAffiliationsAndNotice{\icmlEqualContribution} 

\begin{abstract}
We provide a new efficient version of the backpropagation algorithm, specialized to the case where the weights of the neural network being trained are \emph{sparse}. 
Our algorithm is general, as it applies to arbitrary (unstructured) sparsity and common layer types (e.g., convolutional or linear). 
We provide a fast vectorized implementation on commodity CPUs, and show that it can yield speedups in end-to-end runtime experiments, both in transfer learning using already-sparsified networks, and in training sparse networks from scratch. Thus, our results provide the first support for sparse training on commodity hardware. 
\end{abstract}

\vspace{-1.5em}
\section{Introduction}

The significant computational costs of deep learning have led to massive interest in approaches for leveraging \emph{sparsity} in neural networks, which have been investigated in great breadth and depth~\cite{hoefler2021sparsity}. On the inference side, there is already emerging algorithmic and system support for sparsity on both GPUs~\cite{NVIDIASparse, gale2020sparse} and CPUs~\cite{elsen2020fast, deepsparse}, as well as a wide range of methods for obtaining models which are both highly-sparse and highly-accurate. 

A new frontier in the area is \emph{accurate and efficient sparse training}. 
On the algorithmic side, there are several interesting proposals for \emph{sparse training} algorithms~\cite{dettmers2019sparse, kusupati2020soft, evci2020rigging, jayakumar2020top, schwarz2021powerpropagation}, 
i.e. variants of stochastic gradient descent (SGD) which aim to keep as many weights as possible sparse during training. 
Another interesting related approach is \emph{sparse transfer}~\cite{ zafrir2021prune, chen2021lottery, iofinova2022well, kurtic2022optimal}, by which models sparsified on a large pretraining corpus are then used for \emph{transfer learning} 
on different tasks, while preserving the sparsity mask. 

Despite this progress on the optimization side, the vast majority of these approaches lack \emph{system support for fast training}, in that they do not provide any practical speedups. 
This is because the weight sparsity introduced is \emph{unstructured}, which is notoriously hard to leverage for computational gains. 
Specifically, there is no general implementation of backpropagation that can leverage unstructured weight sparsity for practical speedup on common hardware. 
At the same time, approaches leveraging \emph{block sparsity}~\cite{NVIDIASparse, gray2017gpu} can only reach lower sparsity without significant accuracy drops, and require specialized training algorithms~\cite{lagunas2021block, jiangexposing}. 
As such, unstructured weight sparsity is often dismissed as a practical way of accelerating model training. 

\paragraph{Contribution.} 
We contradict this conventional wisdom by presenting a new vectorized implementation of backpropagation~\cite{rumelhart1986learning}, designed to be efficient in the case where the weights of the neural network are \emph{sparse}, i.e. contain a significant fraction of zero values, and show its potential for practical speedups in common edge training scenarios, for both vision and language tasks.  

More precisely, our algorithm, called SparseProp, is general in the sense that 1) it applies to arbitrary sparsity patterns,  
2) general layer types, and 3) can be efficiently vectorized using standard CPU-supported approaches. 
The asymptotic complexity of the algorithm is \emph{linear} in the \emph{layer density}, i.e. the number of non-zero weights in the layer, providing proportional runtime improvements to the weight sparsity, for both linear and convolutional layers. 

To illustrate practical efficiency, we provide a fast vectorized implementation of SparseProp aimed at general-purpose Intel and AMD CPU architectures. Specifically, our implementation provides drop-in replacement implementations for standard layer types, and only relies on widely-supported  AVX2 instructions. 
We show that SparseProp can lead to practical runtime improvements both on single sparse layers, validating our linear sparsity scaling claims, as well as on end-to-end training of sparse models. 

Specifically, we provide results for a preliminary integration with Pytorch~\cite{pytorch}, 
which can run sparse backpropagation for linear and convolutional layers, covering most popular model families. 
As such, SparseProp can provide direct support for methods like Gradual Pruning~\cite{zhu2017prune}, RigL~\cite{evci2020rigging} or AC/DC~\cite{peste2021ac}, which assume a fixed sparsity mask for any fixed forward and backward pass, and can be modified to support more complex methods~\cite{jayakumar2020top}, 
which specify different sparsities for weights and gradients. 
We believe it is the first implementation to do so on commodity hardware. 

Our end-to-end experiments aim to make the case 
that sparsity can be a viable option for DNN training \emph{at the edge}.
That is, we explore settings where a device with moderate computational power (e.g., a CPU with a limited number of cores) performs either \emph{sparse transfer} or \emph{from-scratch sparse training} over a specialized task. 
We investigate sparsity-versus-accuracy trade-offs in two model/task combinations: 
1) ResNets~\cite{he2016deep} applied to twelve popular vision tasks~\cite{kornblith2019better}, and 2) a standard BERT-base model~\cite{devlin2018bert} applied to GLUE language modelling tasks~\cite{wang2018glue}. 

In the \emph{sparse transfer} scenario, we are provided an already-sparse model pretrained on a large corpus, e.g. ImageNet~\cite{imagenet} respectively WikiText~\cite{wikitext103}, 
and wish to finetune the corresponding sparse weights on a (usually smaller) target dataset. 
This application has gained significant popularity~\cite{zafrir2021prune, chen2021lottery, iofinova2022well, kurtic2022optimal}, and pretrained sparse models are available for several standard tasks and models~\cite{wolf2019huggingface, sparsezoo}. 
In this context, we show that, for both vision and language tasks, 
SparseProp can lead to end-to-end sparse transfer speedups of up to 1.85x, 
at similar accuracies, relative to CPU-based finetuning of \emph{dense models} in the same environment. 
Measured only over backward-pass operations---and thus omitting framework-level overheads--our algorithm provides speedups of 3.6x at 95\% model sparsity. 

In the second scenario, we examine the ability of SparseProp to provide speedups for \emph{sparse training from scratch}, 
on the same series of tasks, adapting variants of sparse training~\cite{zhu2017prune} 
to our setting. 
Experiments show that, in this scenario, SparseProp leads to end-to-end speedups of up to 1.4x, with moderate accuracy loss. 

In sum, our results show that SparseProp can efficiently provide system support for CPU-based unstructured sparse training, ensuring speedups for both from-scratch training and sparse transfer. 
We believe our approach could lead to additional practical impact for research on sparsity, especially given that our end-to-end runtime numbers can still be improved via additional optimizations, and by mitigating external, framework-specific overheads.

\section{Related Work}

\paragraph{Sparse Inference.} 
One of the key motivations behind sparsity in DNNs is reducing inference costs. 
For this, an impressive number of weight pruning techniques have been introduced, e.g.~\cite{lecun1990optimal, hagiwara1994, han2015deep, singh2020woodfisher, 2020-sanh}. 
Complementing this work, there have been a number of algorithmic proposals for efficient sparse inference algorithms over DNNs, e.g.~\cite{park2016faster, han2016eie, gale2020sparse, elsen2020fast}, although it is known that layer-wise gains can be difficult to translate into end-to-end speedups~\cite{wang2020sparsert}.  
Nevertheless, sparse inference support is now available on both CPUs, e.g.~\cite{deepsparse} and GPUs~\cite{NVIDIASparse}.

\citet{hubara2021accelerated} proposed a theoretically-justified approach for identifying sparse transposable masks matching the NVIDIA 2:4 sparsity pattern, which could be leveraged for faster training on GPUs. 
However, they do not provide an implementation, and, currently, GPU-based 2:4 sparsity speedups tend to be minimal~\cite{NVIDIA24}.

\paragraph{Sparse SGD-Based Training.} 
As noted, there has been a significant amount of work on SGD-like algorithms for \emph{sparse training} of DNNs, 
balancing accuracy while trying to maximize sparsity in the models' internal representations~\cite{mocanu2016topological, bellec2017deep, zhu2017prune, mostafa2019parameter, lis2019full, dettmers2019sparse, zhang2020memorized, wiedemann2020dithered, kusupati2020soft, evci2020rigging, jayakumar2020top, peste2021ac, schwarz2021powerpropagation}. 
Unfortunately, a precise comparison is quite difficult, since each makes different assumptions regarding the degree of sparsity in the network's internal representations, potentially even varying the amount of sparsity between weights and gradients, e.g.~\cite{jayakumar2020top}.

\paragraph{Sparse Training for Speedup.} 
Leveraging sparsity for practical speedups has been a major goal in model compression~\cite{hoefler2021sparsity}. 
\citet{yang2020procrustes} proposed a specialized hardware accelerator which is specialized to the DropBack pruning algorithm~\cite{lis2019full}. 
SWAT~\citep{raihan2020sparse} proposed a sparsity-aware algorithm for the case where both weights and activations have high sparsity, and showed speedups in a simulated environment. 
Their approach works only for specific networks, and can lose significant accuracy.  
More recently,~\citet{jiangexposing} proposed an algorithm-hardware co-design approach, for the case of GPU-based training. Specifically, their approach imposes \emph{block sparsity} in GPU-friendly patterns, and leverages it for speedup.  

By contrast to this work, our approach considers efficient support for backpropagation for \emph{unstructured sparse weights}, implements this efficiently for commodity CPUs, and shows that this can be leveraged for \emph{end-to-end speedup} during training. 
Specifically, this provides support to the vast amount of existing work on unstructured sparse training algorithms, 
on commodity hardware. 

\paragraph{System Support.}  
Pytorch~\cite{paszke2019pytorch} introduced partial sparse tensor support recently, while the STen~\cite{ivanov2022sten} provides a general interface for such representations. 
Our work is complementary to this direction, as our implementation can be  interfaced with Pytorch or specifically STen to provide training speedups.

\section{The Sparse Backpropagation Algorithm}

\subsection{Background}

\paragraph{SIMD Instructions.}
Fast and efficient numerical code heavily relies on Single Instruction Multiple Data (SIMD) instructions to improve performance. These instructions operate on specialized machine registers ($\texttt{xmm}$, $\texttt{ymm}$, and $\texttt{zmm}$) that contain multiple values. 

Our implementations currently support $\texttt{x86}$ machines that provide the standard AVX2 instruction set, which uses $256$ bit registers, or $8$ single precision floating point values. Table~\ref{tbl:instructions} provides an overview of the instructions employed by our library. Our SIMD implementation structure follows the Load-Compute-Store paradigm, where data is explicitly transferred to registers via the \avx{loadv} and \avx{broadcastv} instructions. Computation is performed on the data in the registers using fused multiply-add instructions \avx{vfmadd} ($r = a \cdot b + c$), and the results are subsequently moved back to memory with the \avx{vstore} instruction. Following this structure, we significantly increase performance since the data loaded into the registers can be used for multiple operations before being stored back in memory.

\paragraph{Backpropagation.} Let $f(\mathbf{X};\mathbf{W})$ represent a layer (fully-connected or convolution) in a neural network $\mathcal{N}$; $\mathbf{W}$ represents the parameters of this layer and $\mathbf{X}$ represents a batch of inputs. Let $B$ be the batch size. Additionally, denote the output of this layer by $\mathbf{O}=f(\mathbf{X};\mathbf{W})$. Let $L$ be the loss of the whole network $\mathcal{N}$ for this batch of inputs. Back-propagating through this layer involves calculating the gradients $\partial L / \partial \mathbf{W}$ and $\partial L / \partial \mathbf{O}$, given $\partial L /\partial \mathbf{O}$.

Consider the situation where we have a highly sparsified matrix $\mathbf{W}$ that is stored as a sparse matrix. During the backpropagation process, it is necessary to calculate the gradient of this matrix. However, in practice, the full gradient of the dense matrix is often calculated, even though the pruned elements are not updated and their gradients are discarded. This can be inefficient, as it consumes a significant amount of computation and time without providing any benefits.

\subsection{The Case of Fully-connected Layers}
We now focus on the case where $f(.)$ is a fully-connected layer. Assume $\mathbf{X}$ and $\mathbf{W}$ are $B \times M$ and $M \times N$ matrices, respectively. Consequently, $\mathbf{O} = f(\mathbf{X};\mathbf{W}) = \mathbf{X} \mathbf{W}$ will be a $B \times N$ matrix. The gradients of $L$ with respect to $\mathbf{X}$ and $\mathbf{W}$ are calculated as follows:

\begin{eqnarray}
\frac{\partial L}{\partial \mathbf{X}} = \frac{\partial L}{\partial \mathbf{O}} \mathbf{W}^T    
\label{fc-back-x}
\\ 
\frac{\partial L}{\partial \mathbf{W}} = \mathbf{X}^T \frac{\partial L}{\partial \mathbf{O}}
\label{fc-back-w}
\end{eqnarray}

If we examine equations \eqref{fc-back-x} and \eqref{fc-back-w}, we can see that the former is a General Sparse Matrix-Matrix Multiplication (SpGEMM) operation, while the latter is a Sampled Dense Dense Matrix Multiplication (SDDMM) operation. 

\paragraph{Sparse Representation.}
The matrix $\mathbf{W}$ is stored in a compressed sparse row (CSR) format, a standard representation for sparse matrices. The non-zero values of the matrix are stored in the arrays $W_{\nnzvals}$ and $W_{\nnzcols}$, which correspond to the values and column indices of the non-zero elements, respectively. The array $W_{\nnzrows}$ encodes each row's start and end indices. For example, the non-zero values and column indices of a row $i$ of $\mathbf{W}$ are contained between positions $W_{\nnzrows}[i]$ and $W_{\nnzrows}[i+1]$.

\begin{table}[t]
\centering
\ra{1.3}
\footnotesize
\begin{tabular}{@{}ll@{}}\toprule
\avx{vload}(\texttt{address}) & load from memory address \\
\avx{vstore}(\texttt{address,a}) & store \texttt{a} at memory address \\
\avx{vbroadcast}(\texttt{a}) & fill a register with \texttt{a}\\
\avx{vfmadd}(\texttt{a,b,c}) & return $a \cdot b + c$\\
\avx{vaddreduce}(\texttt{a}) & return sum elements of \texttt{a}\\
\bottomrule
\end{tabular}
    \caption{List of vector instructions used in the implementation and their semantics.}\label{tbl:instructions}
\end{table}

\paragraph{Algorithm.}
In Algorithm~\ref{alg:lin-backprop}, we present high-level pseudocode for backpropagation in our linear layer. The calculations for \eqref{fc-back-x} and \eqref{fc-back-w} are performed in a single pass by utilizing the sparsity pattern of $\mathbf{W}$, which is identical to $\partial L / \partial \mathbf{W}$. Specifically, the result of $(\partial L / \partial \mathbf{O}) \mathbf{W}^T$ is computed as a sparse matrix-matrix multiplication. Whereas  $\mathbf{X}^T (\partial L / \partial \mathbf{O})$ is computed as an SDDMM, with $\nnz$ dot-products, where $\nnz$ is the number of non-zero elements of $\mathbf{W}$. In more detail, the computation is divided into $3$ loops. The innermost loop contains the core of the computation. It computes at each iteration $16$ floating point operations using $2$ \avx{fmadd} instructions: the first \avx{fmadd} computes $8$ entries of  $\partial L / \partial \mathbf{X}$ and the second accumulate a dot-product in a register \texttt{acc}. 

\begin{figure}
    \centering
    \includegraphics[scale=0.88]{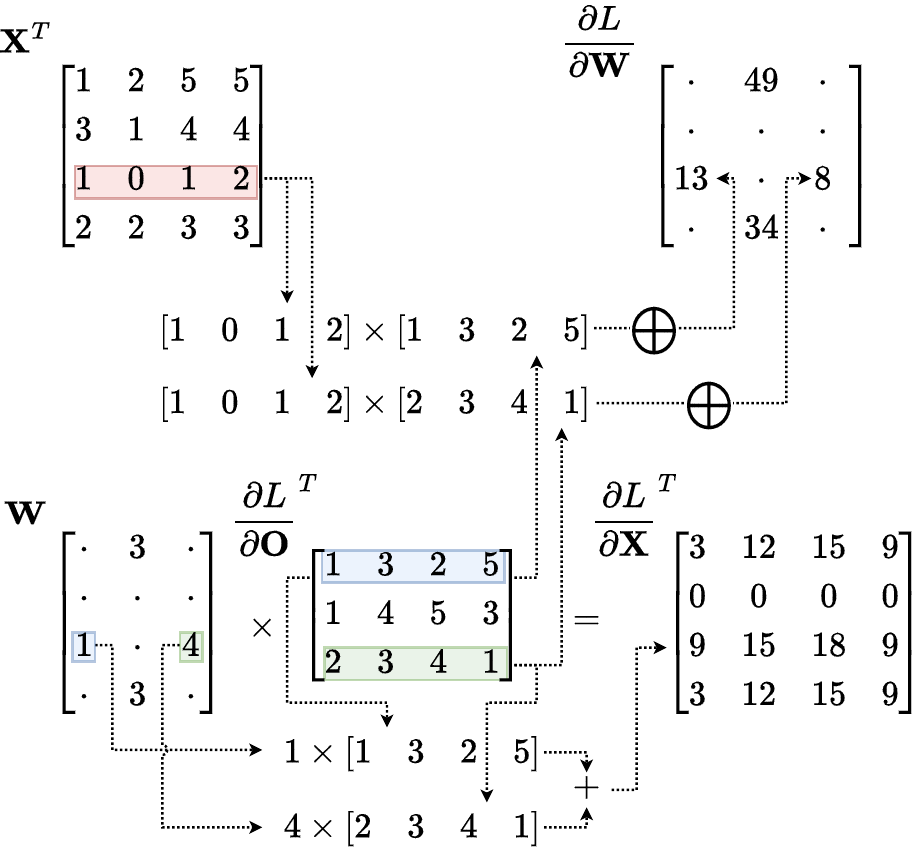}
    \caption{Visual representation of the core computation of Algorithm~\ref{alg:lin-backprop} using vector registers of size $4$. We represent elementwise multiplication with $\times$ and \avx{vaddreduce} with $\bigoplus$.}
    \label{fig:SDDMM}
\end{figure}

\paragraph{Implementation Details.}
We operate on the transposed matrices in our linear layer implementation to improve cache utilization. Specifically, in both the forward and backward passes, we operate on the transposed version of the input matrix, $\mathbf{X}^T$, which is a column-major representation of $\mathbf{X}$. By doing so, we achieve a streaming access pattern over the rows of $\mathbf{X}^T$, as we see from the innermost loop Algorithm~\ref{alg:lin-backprop}. Additionally, we leverage that the transpose of a CSR matrix is none other than the same matrix in Compressed-Sparse-Column (CSC) format to avoid expensive sparse transpose operations. An example computation is given in Figure~\ref{fig:SDDMM}.

\begin{algorithm}[tb]
 \footnotesize
   \caption{AVX2 Linear Backward Pass}
   \label{alg:lin-backprop}
\begin{algorithmic}
\renewcommand\algorithmiccomment[1]{%
  /* {#1} */%
}
    
    \FOR{$i=0$ {\bfseries to} $N-1$}
        \FOR{$j=\mathbf{W}_{\nnzcols} [i]$ {\bfseries to} $\mathbf{W}_{\nnzcols} [i+1]$}
            \STATE \COMMENT{\texttt{repeat one weight entry 8 times}}
            \STATE $v \leftarrow \avx{vbroadcast}(\mathbf{W}_{\nnzvals} [j])$
            \STATE \COMMENT{\texttt{initialize} $acc$ \texttt{to zero}}
            \STATE $acc \leftarrow \avx{vbroadcast}(0)$
            \STATE $r \leftarrow \mathbf{W}_{\nnzrows} [j]$
            \FOR{$k=0$ {\bfseries to} $B-1$ {\bfseries with} $k\mathrel{{+}{=}}8$}
                \STATE \COMMENT{\texttt{load 8 values}}
                \STATE $dx \leftarrow \avx{vload}((\partial L / \partial \mathbf{X})_{r,k})$
                \STATE $do \leftarrow \avx{vload}((\partial L / \partial \mathbf{O})_{r,k})$
                \STATE $x \leftarrow \avx{vload}(\mathbf{X}_{i,k})$
                \STATE \COMMENT{\texttt{compute} $8$ $dx = do \cdot v + dx$}
                \STATE $dx \leftarrow \avx{vfmadd}(do, v, dx)$
                \STATE \COMMENT{\texttt{compute} $8$ $acc = do \cdot x + acc$}
                \STATE $acc \leftarrow \avx{vfmadd}(do, x, acc)$
                \STATE \COMMENT{\texttt{store updated} $dx$ \texttt{back}}
                \STATE $\avx{vstore}((\partial L / \partial \mathbf{X})_{r,k}, dx)$
            \ENDFOR
            \STATE \COMMENT{\texttt{sum the} $8$ \texttt{values in} $acc$}
            \STATE $(\partial L / \partial \mathbf{W})_{\nnzvals}[j] \leftarrow \avx{vaddreduce}(acc)$
        \ENDFOR
   \ENDFOR
\end{algorithmic}
\end{algorithm}

\subsection{Sparse Backpropagation for Convolutional Layers}
We now examine convolutional layers. Denote the number of input channels by $IC$, and the number of output channels by $OC$. The input width and height are represented by $M$ and $N$, respectively, and let the kernel size be $K \times K$. The input tensor, $\mathbf{X}$, and the weights tensor, $\mathbf{W}$, have dimensions $B \times IC \times M \times N$ and $OC \times IC \times K \times K$, respectively. For simplicity, here we only consider the case where padding is $0$ and stride is $1$. For larger padding and stride, the generalization is not difficult. The output tensor, $\mathbf{O}$, will be of size $B \times OC \times OM \times ON$, where $OM = M - K + 1$ and $ON = N - K + 1$. For $0 \le b < B, 0 \le oc < OC, 0 \le p < OM, 0 \le q < ON$ we have:
\begin{equation}
    \begin{aligned}
    \mathbf{O}[b, oc, p, q] & = \sum_{ic=0}^{IC-1} \sum_{i=0}^{K-1} \sum_{j=0}^{K-1} \mathbf{W}[oc, ic, i, j] \\
    & ~~~~~~~~~~~~~~~~~~ . \mathbf{X}[b, ic, p+i, q+j].
    \end{aligned}
\end{equation}
Using the chain rule, it is easy to check that for $0 \le b < B, 0 \le ic < IC, 0 \le m < M, 0 \le n < N$:
\begin{equation}
    \label{conv-back-x}
    \begin{aligned}
        \frac{\partial L}{\partial \mathbf{X}}[b, ic, m, n] & = \sum_{oc=0}^{OC-1} \sum_{p=p_s}^{m} \sum_{q=q_s}^{n} \frac{\partial L}{\partial \mathbf{O}}[b,oc,p,q] \\
        & ~~~~~~~~~~~~~~~ . \mathbf{W}[oc, ic, m-p, n-q],
    \end{aligned}
\end{equation}
with $p_s=m-K+1$, and $q_s=n-K+1$. 
And for $0 \le oc < OC, 0 \le ic < IC, 0 \le i < K, 0 \le j < K$, we have:
\begin{equation}
    \label{conv-back-w}
    \begin{aligned}
        \frac{\partial L}{\partial \mathbf{W}}[oc, ic, i, j] & = \sum_{b=0}^{B-1} \sum_{p=0}^{M-K} \sum_{q=0}^{N-K} \frac{\partial L}{\partial \mathbf{O}}[b,oc,p,q] \\
        & ~~~~~~~~~~~~~~~~~~~~~ . \mathbf{X}[b, ic, p+i, q+j].
    \end{aligned}
\end{equation}

It is assumed that the weight matrix $\mathbf{W}$ is sparse. In accordance with equation (\ref{conv-back-x}), when a weight $\mathbf{W}[oc, ic, m-p, n-q]$ is pruned, the multiplication and corresponding addition operations can be skipped. Furthermore, when a weight $\mathbf{W}[oc, ic, i, j]$ is pruned, the calculation of the gradient for this parameter is not necessary, as it will not be updated, and therefore the computation outlined in equation (\ref{conv-back-w}) can be skipped.

\paragraph{Sparse Representation.}
To efficiently represent a sparse tensor, we employ a representation akin to the compressed sparse row (CSR) format used for sparse matrices. Four arrays, $W_{\nnzoc}$, $W_{\nnzic}$, $W_{\nnzx}$, and $W_{\nnzy}$, are used to store the indices, and an array $W_{\nnzvals}$ is used to store the non-zero values. Specifically, $W_{\nnzx}$ and $W_{\nnzy}$ are arrays of size $\nnz$, which contain the coordinates of the non-zero values of each filter. $W_{\nnzoc}$ is an array of size $OC + 1$, which encodes the start of each output channel's entries in $W_{\nnzic}$. Finally, $W_{\nnzic}$ is an array of size $OC \times (IC + 1)$, which encodes the indices in $W_{\nnzx}$, $W_{\nnzy}$, and $W_{\nnzvals}$ of each input channel. For example, for an output channel $oc$, the non-zero elements of the input channel $ic$ are stored between indices $W_{\nnzoc}[oc] + W_{\nnzic}[oc \cdot (IC + 1) + ic]$ and $W_{\nnzoc}[oc] + W_{\nnzic}[oc \cdot (IC + 1) + ic+1]$. As example, consider the following sparse tensor of dimensions $(3,2,2,3)$

\begin{displaymath}
\begin{Bmatrix}
\begin{bmatrix}
\zero&a&\zero\\
\zero&\zero&\zero
\end{bmatrix}&\begin{bmatrix}
\zero&\zero&\zero\\
\zero&\zero&\zero
\end{bmatrix}&\begin{bmatrix}
\zero&\zero&\zero\\
b&\zero&c
\end{bmatrix}\\
&\\
\begin{bmatrix}
\zero&\zero&\zero\\
\zero&\zero&\zero
\end{bmatrix}&\begin{bmatrix}
\zero&d&\zero\\
\zero&e&\zero
\end{bmatrix}&\begin{bmatrix}
\zero&\zero&f\\
\zero&\zero&\zero
\end{bmatrix}
\end{Bmatrix},
\end{displaymath}
where $\zero$ represents a zero value. Its sparse representation is given by
\begin{align*}
W_{\nnzoc} &= \begin{pmatrix} 0 & 1 & 3 & 6 \end{pmatrix},\\
W_{\nnzic} &= \begin{pmatrix} 0& 1& 1& 0& 0& 2& 0& 2& 3 \end{pmatrix},\\
W_{\nnzx} &= \begin{pmatrix} 0& 0& 1& 1& 1& 0 \end{pmatrix},\\
W_{\nnzy} &= \begin{pmatrix} 1& 1& 1& 0& 2& 2 \end{pmatrix},\\
W_{\nnzvals} &= \begin{pmatrix} a & b & c & d & e & f \end{pmatrix}.
\end{align*}
Although the usage of a $W_{\nnzoc}$ may seem superfluous, its inclusion allows us to reduce total memory usage. Indeed, assuming $K < 256$ and $IC < 8192$, we can store entries in $W_{\nnzx}$ and $W_{\nnzy}$ using \texttt{uint8\_t} and 
$W_{\nnzic}$ using \texttt{int16\_t}. Therefore, using $W_{\nnzoc}$ lowers memory usage from $4B \times (IC + 1) \times OC$ bytes to $2B\times (IC + 1) \times OC + 4 \times OC$ bytes.

\paragraph{Algorithm.}
In Algorithm~\ref{alg:conv-backprop}, we present an overview of our backpropagation algorithm for a convolutional layer. If we ignore at first the pointer arithmetic needed to traverse the structures, the main structure remains similar to that of Algorithm~\ref{alg:lin-backprop} as both innermost loops use the same instructions.

\paragraph{Implementation Details.}
We developed two kernels for fast 2D convolutions based on the input dimensions. For larger $M$ and $N$, we found that no preprocessing was needed. Keeping the input dimensions as $B \times IC \times M \times N$  offers both computing batches in parallel on multiple threads and high single-core parallelization using AVX2 instructions. On the other hand, for small $M$ and $N$, which often occur for the last layers of a network, we found it more efficient to permute the tensors to have the batch as the last index. In particular, $\mathbf{X}$ and $\partial L / \partial \mathbf{X}$ became of size $IC \times M \times N \times B$, and $\mathbf{O}$ and  $\partial L / \partial \mathbf{O}$ became of size $OC \times OM \times ON \times B$. Indeed, setting $B$ as the last dimension allows the usage of SIMD instructions even for small $M$ and $N$.

\begin{algorithm}[tb]
\footnotesize

   \caption{AVX2 Convolutional Backward Pass}
   \label{alg:conv-backprop}
\begin{algorithmic}    
\renewcommand\algorithmiccomment[1]{%
  /* {#1} */%
}
    \FOR{$ic=0$ {\bfseries to} $IC$}
        \FOR{$oc=0$ {\bfseries to} $OC$}
            \STATE $si_s = \mathbf{W}_{\nnzoc}[oc] + \mathbf{W}_{\nnzic}[ic]$
            \STATE $si_e = \mathbf{W}_{\nnzoc}[oc] + \mathbf{W}_{\nnzic}[ic+1]$
            \FOR{$si=si_s$ {\bfseries to} $si_e$}

                \STATE \COMMENT{\texttt{repeat 8 times}}
                \STATE $v \leftarrow \avx{vbroadcast}(\mathbf{W}_{\nnzvals} [j])$
                \STATE \COMMENT{\texttt{initialize} $acc$ \texttt{to zero}}
                \STATE $acc \leftarrow \avx{vbroadcast}(0)$

                \STATE $p_s \leftarrow \max(0, pad - \mathbf{W}_{\nnzx}[si])$
                \STATE $p_e \leftarrow \min(pad - \mathbf{W}_{\nnzx}[si] + M, OM)$
                \STATE $q_s \leftarrow \max(0, pad - \mathbf{W}_{\nnzy}[si])$
                \STATE $q_e \leftarrow \min(pad - \mathbf{W}_{\nnzy}[si] + N, ON)$
                
                \FOR{$p=p_s$ {\bfseries to} $p_e$}
                    \FOR{$q=q_s$ {\bfseries to} $q_e$}
                        \FOR{$k=0$ {\bfseries to} $B-1$ {\bfseries with} $k\mathrel{{+}{=}}8$}
                            \STATE \COMMENT{\texttt{load 8 values}}
                            \STATE $do \leftarrow \avx{vload}((\partial L / \partial \mathbf{O})_{oc,p,q,k})$
                            \STATE $dx \leftarrow \avx{vload}((\partial L / \partial \mathbf{X})_{ic,p,q,k})$
                            \STATE $x \leftarrow \avx{vload}(\mathbf{X}_{ic,p,q,k})$

                            \STATE \COMMENT{\texttt{compute} $8$ $dx = do \cdot v + dx$}
                            \STATE $dx \leftarrow \avx{vfmadd}(do, v, dx)$
                            
                            \STATE \COMMENT{\texttt{compute} $8$ $acc = do \cdot x + acc$}
                            \STATE $acc \leftarrow \avx{vfmadd}(do, x, acc)$

                            \STATE \COMMENT{\texttt{store the updated} $dx$}
                            \STATE $\avx{vstore}((\partial L / \partial \mathbf{X})_{ic,p,q,k}, dx)$
                        \ENDFOR
                    \ENDFOR
                \ENDFOR
            \ENDFOR
            \STATE \COMMENT{\texttt{sum the} $8$ \texttt{values in} $acc$}
            \STATE $(\partial L / \partial \mathbf{W})_{\nnzvals}(j) \leftarrow \avx{vaddreduce}(acc)$
        \ENDFOR
   \ENDFOR
\end{algorithmic}
\end{algorithm}

\section{Experiments}

\paragraph{Setup and Goals.} 
We now experimentally validate our approach. 
First, we perform an in-depth exploration of our algorithm's runtime relative to weight sparsity in a synthetic scenario, i.e. for standard layer and input shapes. 
Then, we examine performance for two \emph{end-to-end training} scenarios, as part of a Pytorch integration. 
Specifically, we examine performance for \emph{sparse transfer}, i.e. fine-tuning of already-sparsified accurate models on a different ``transfer'' dataset, and \emph{from-scratch sparse training} on some specialized tasks. 

\paragraph{Pytorch Modules.}
We provide two Pytorch modules, one for linear and one for convolution layers, facilitating the integration of our algorithm to different applications. To this end, we employ the pybind11 library~\cite{pybind11}  for fast communication between Pytorch and the \CC~backend.

\subsection{Synthetic Performance Evaluation}
\begin{figure}
    \centering
    \subfigure[Linear backward]{
    \includegraphics[scale=.31]{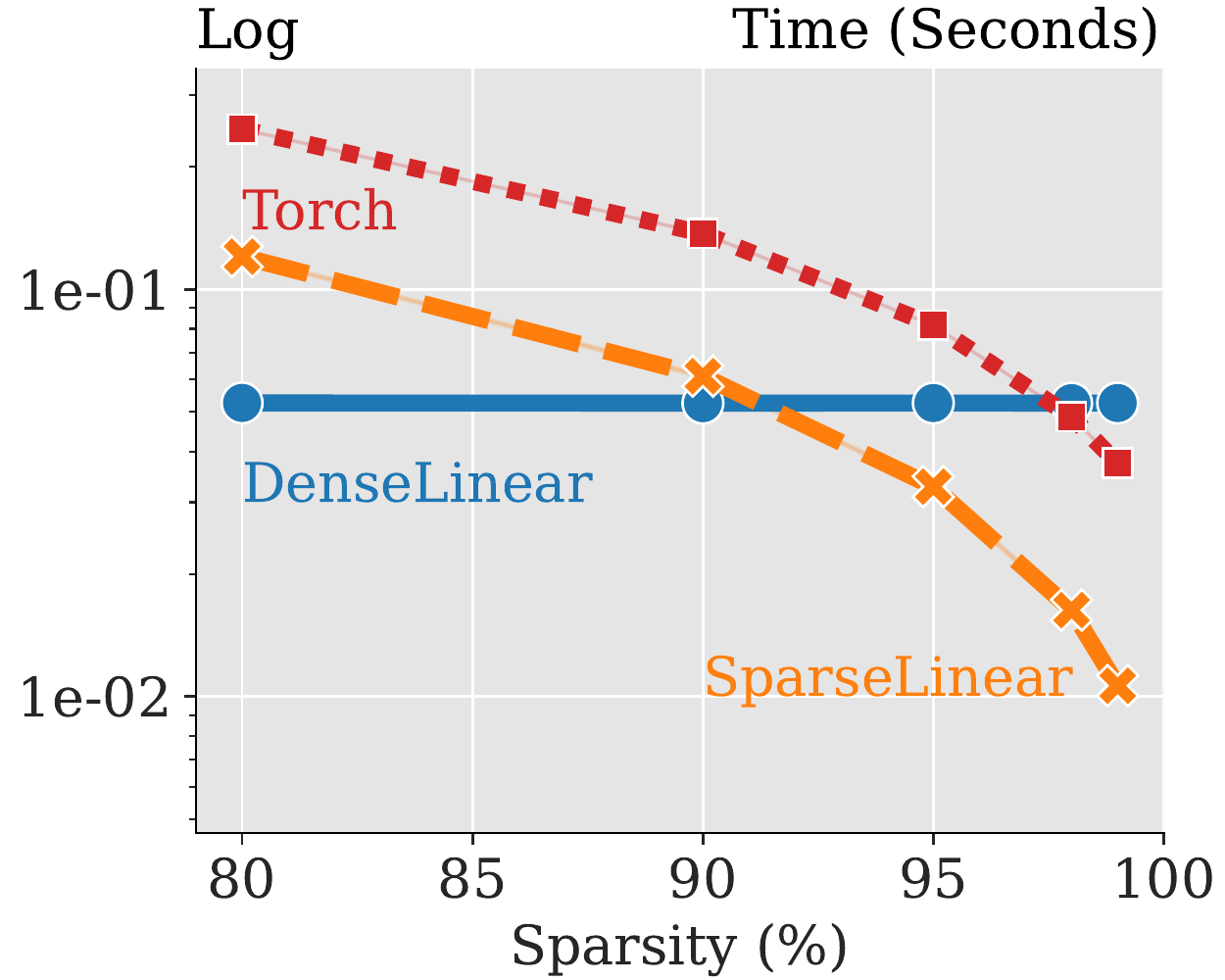}%
    \label{subfig:lin-single-run}%
    }
    \subfigure[Convolutional backward]{
    \includegraphics[scale=.31]{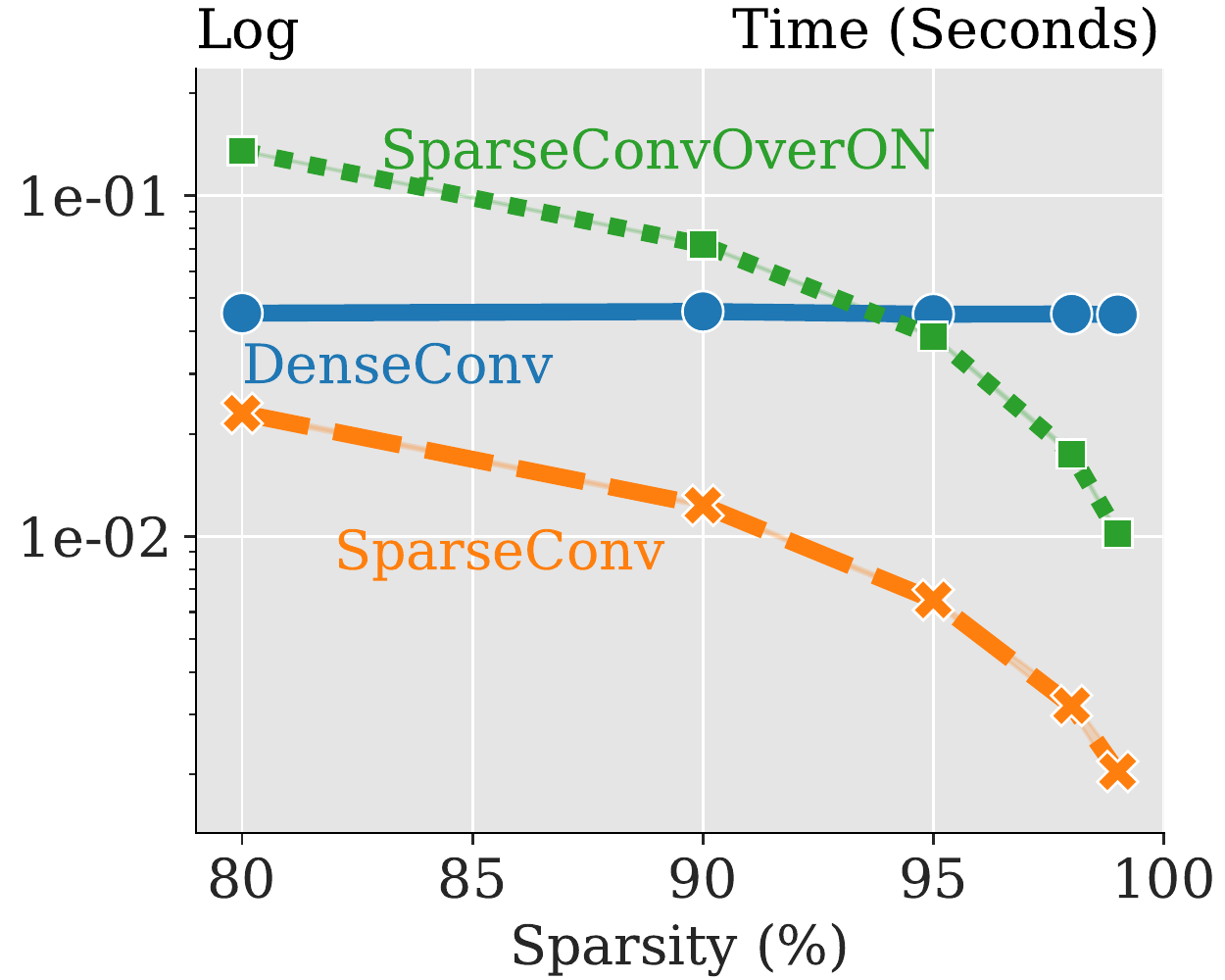}%
    \label{subfig:conv-single-run}%
    }
    \caption{Runtime of our single-core backward propagation of sparse linear and convolutional layers. The former has dimension $(M,N) = (768,3072)$ and the input is of size $(B,M) = (902, 768)$, while the latter is of size $(OC,IC,K,K) = (256,128,3,3)$, and the input is $(B,IC,M,N) = (8,128,7,7)$.}
    \label{fig:single-run}
\end{figure}

\begin{figure}
    \centering
    \subfigure[Parallel linear forward]{
    \includegraphics[scale=.31]{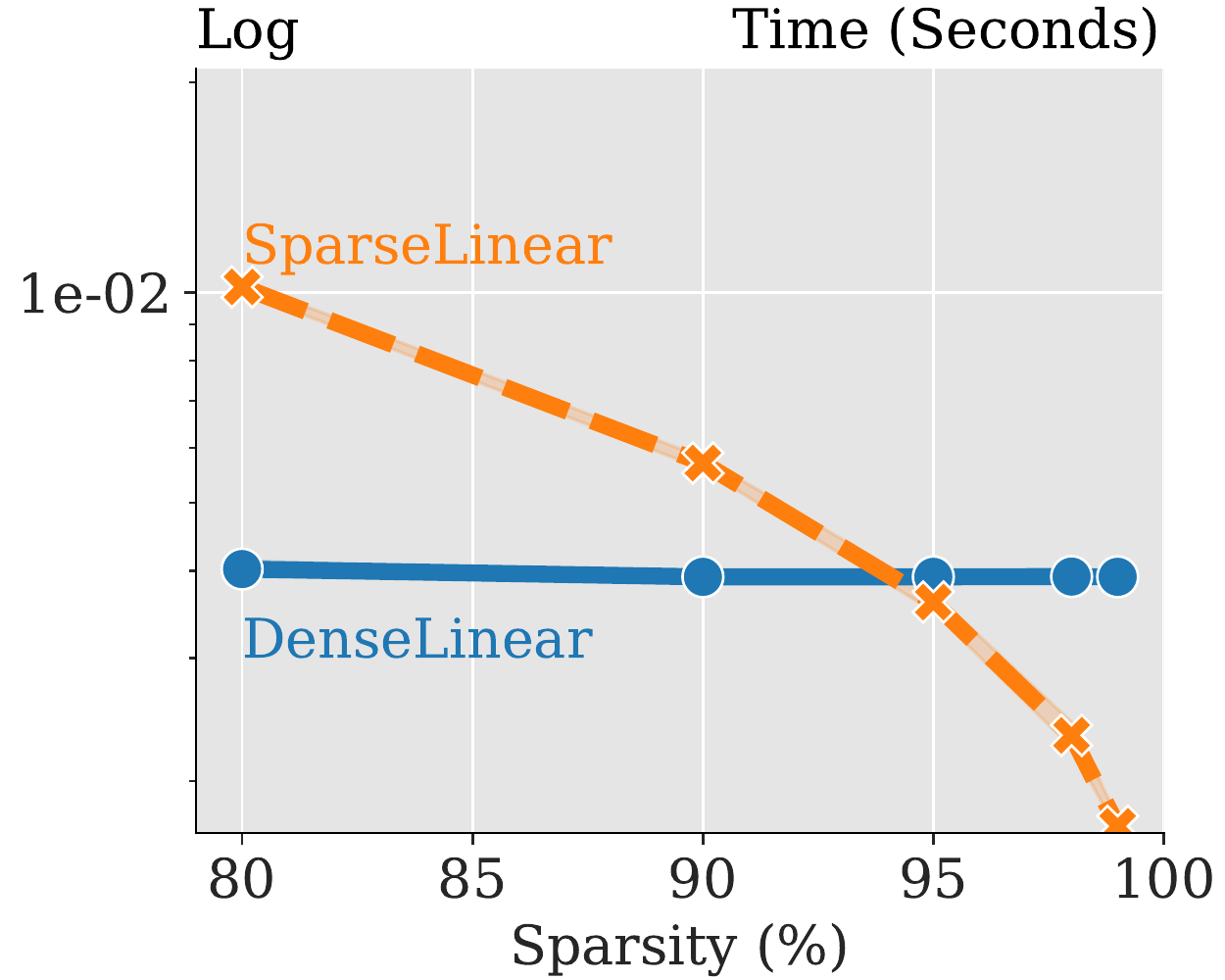}%
    }
    \subfigure[Parallel linear backward]{
    \includegraphics[scale=.31]{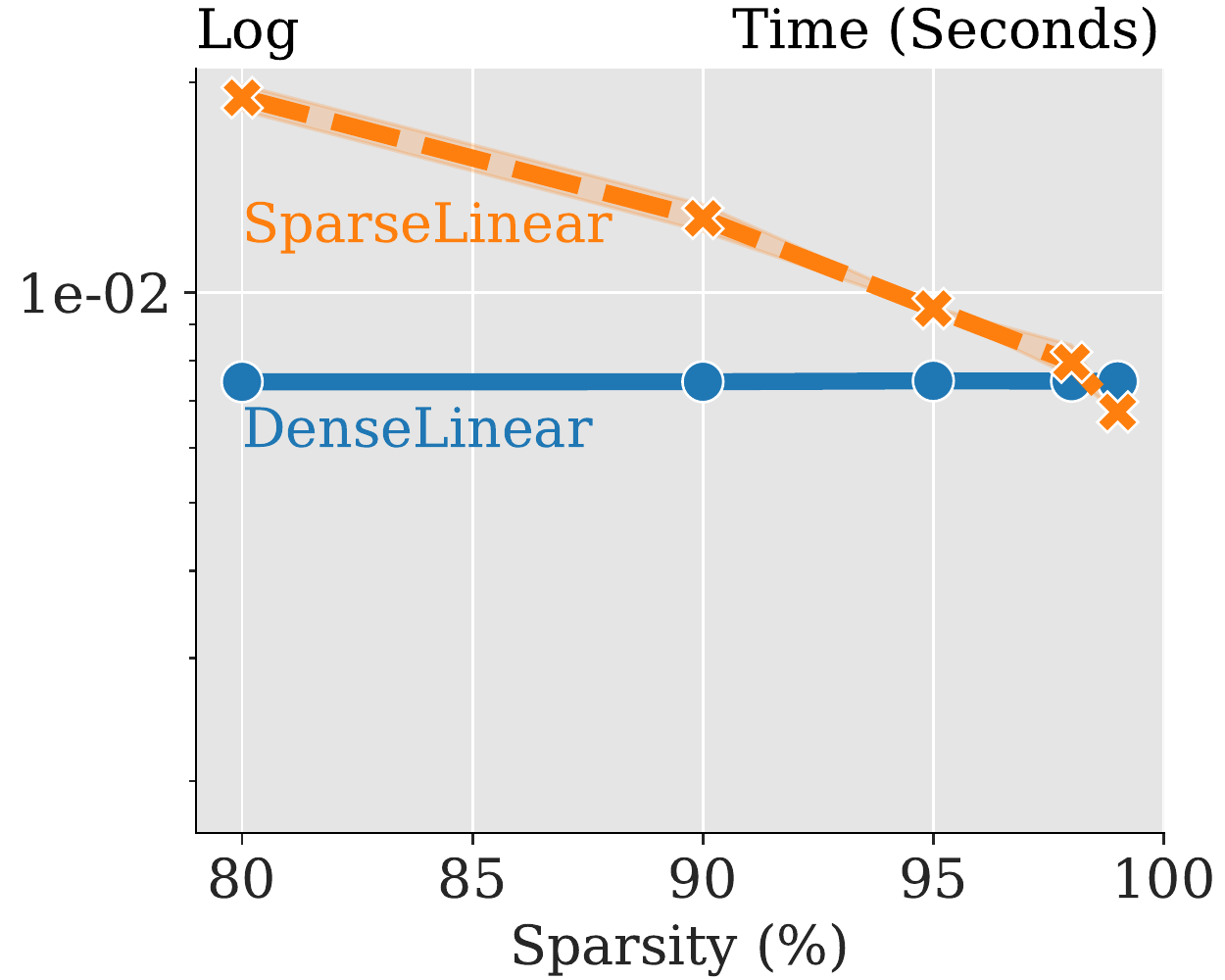}%
    }
    \caption{Runtime measurement of our parallel sparse linear layer compared against a dense implementation. The layer has dimension $(M,N) = (768,3072)$ the input is of size $(B,M) = (902, 768)$.}
    \label{fig:lin-par-run}
\end{figure}

\begin{figure}
\begin{center}
\subfigure[Parallel Conv. forward]{
    \includegraphics[scale=.31]{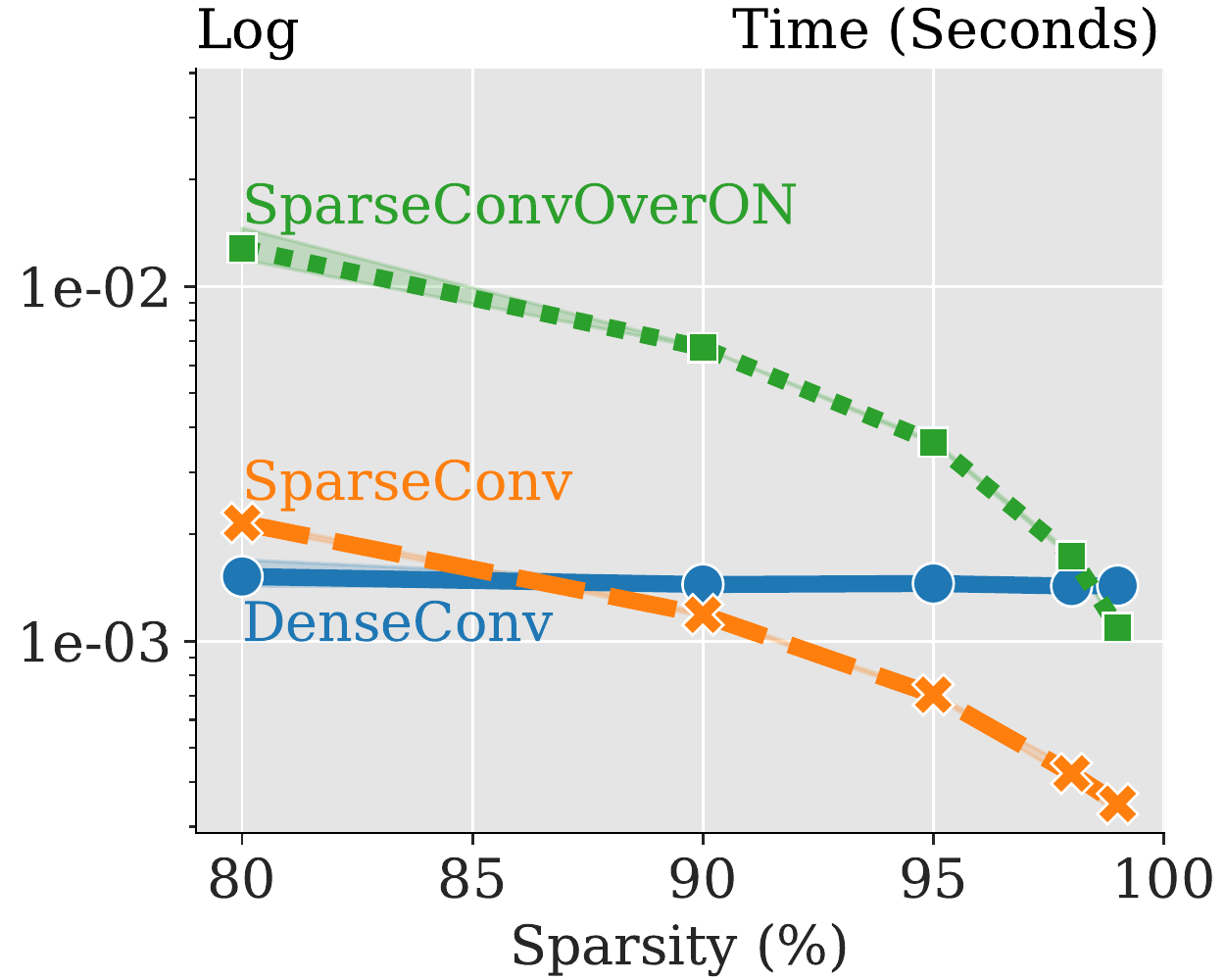}\label{subfig:par-a}%
}
\subfigure[Parallel Conv. backward]{
    \includegraphics[scale=0.31]{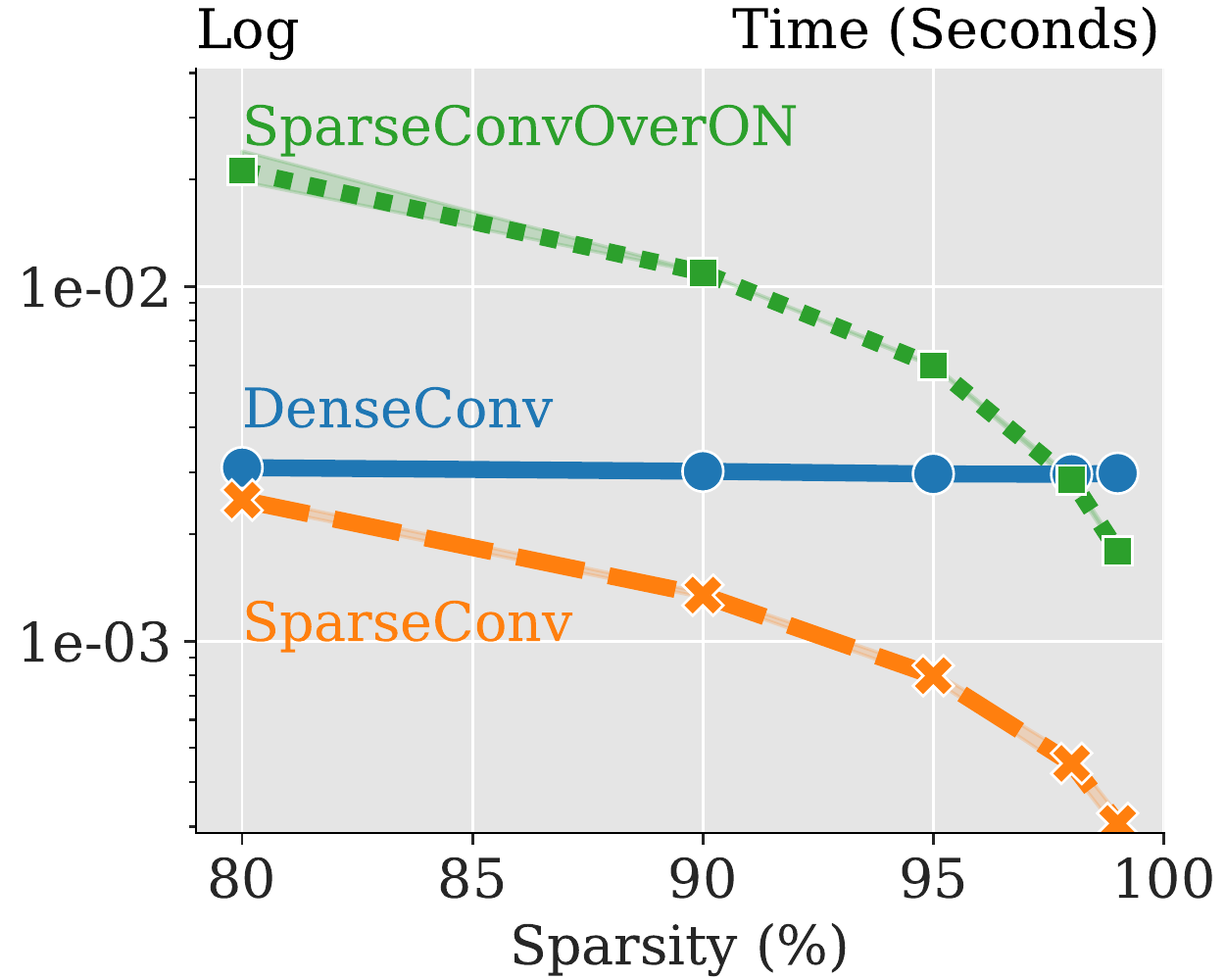}\label{subfig:par-b}%
}
\\
\subfigure[Parallel Conv. forward]{
    \includegraphics[scale=.31]{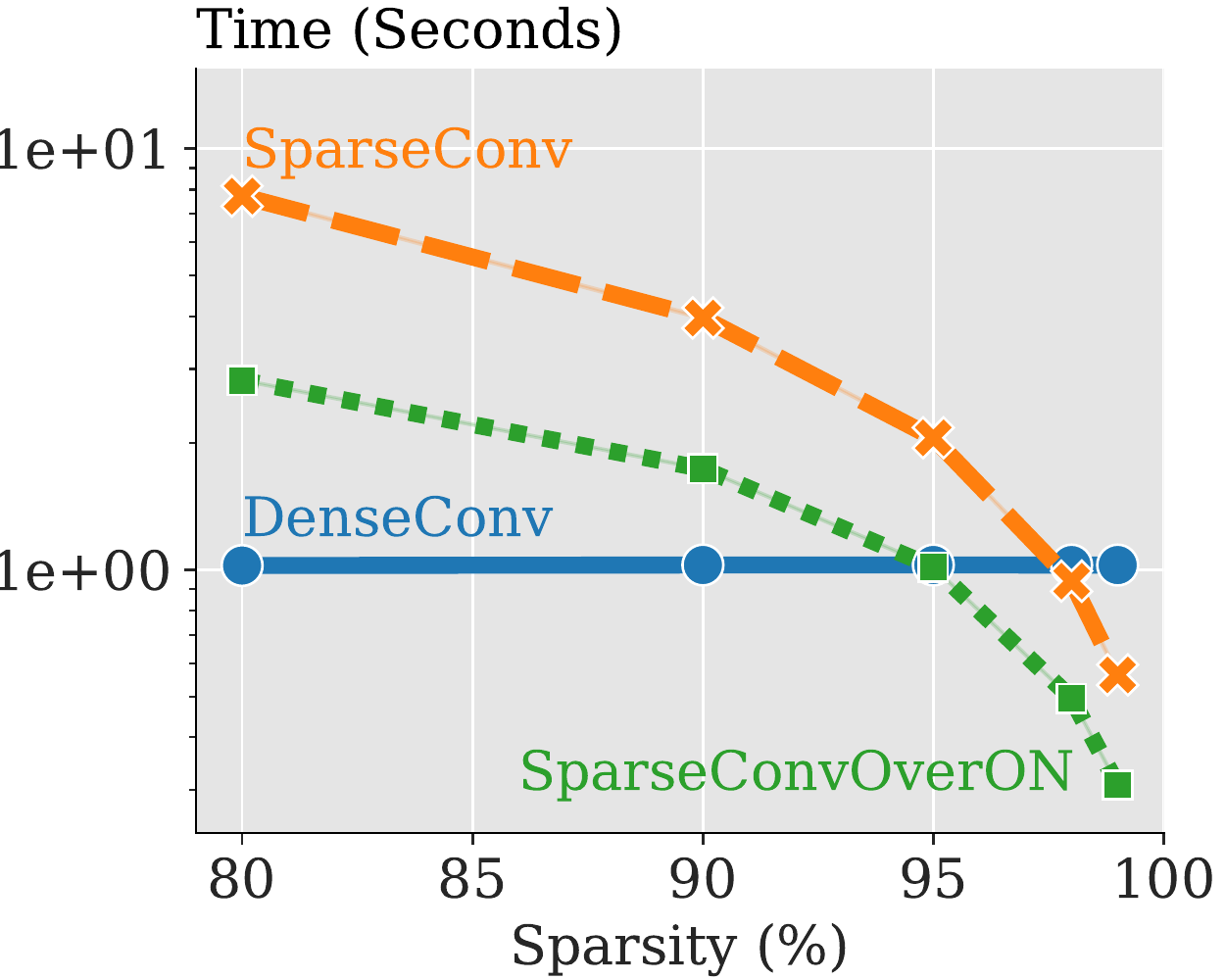}\label{subfig:par-c}%
}
\subfigure[Parallel Conv. backward]{
    \includegraphics[scale=.31]{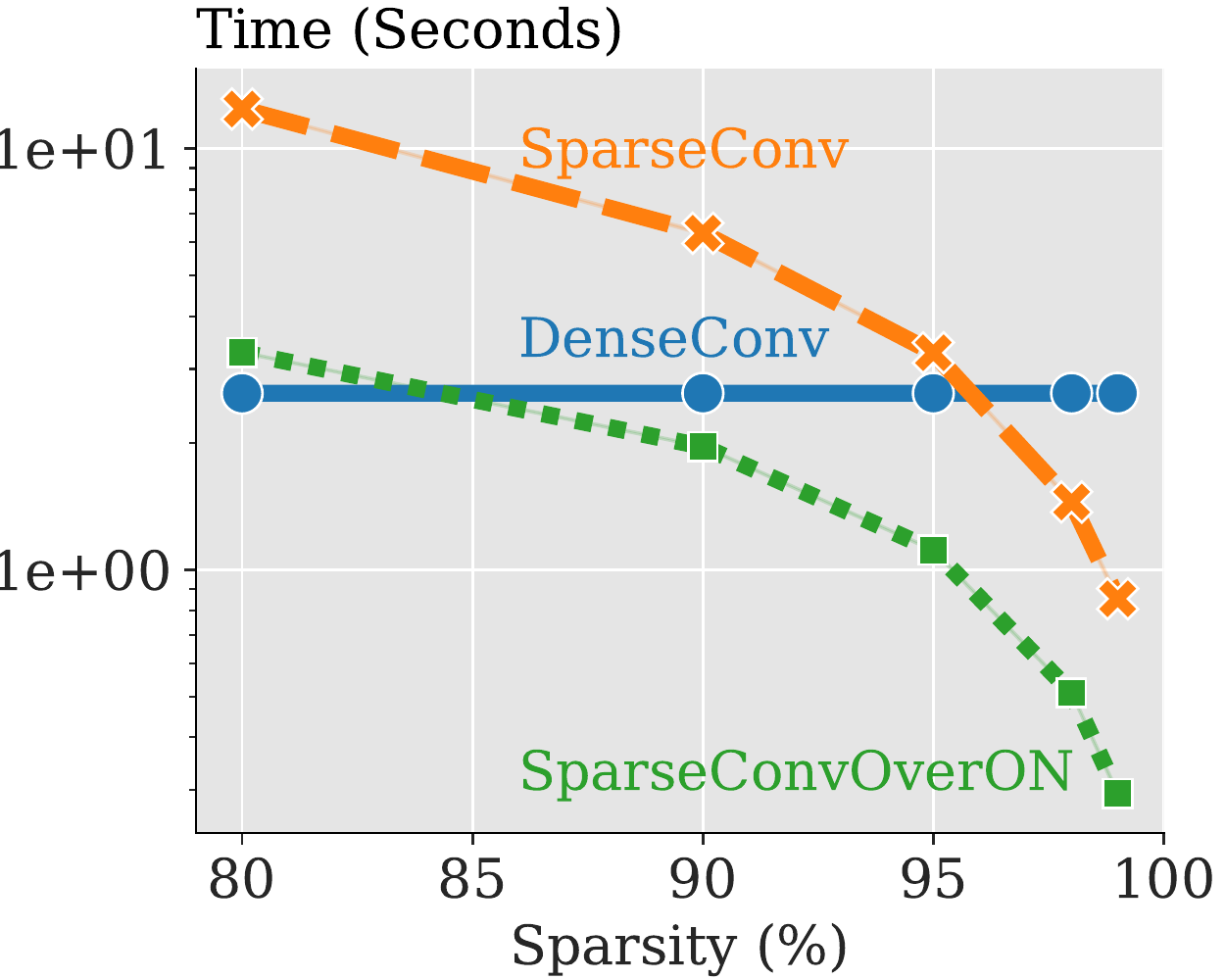}\label{subfig:par-d}%
}
\caption{Runtime measurement of our parallel sparse convolutional layers (base and vectorized over the $ON$ dimension) on $8$ threads compared against a dense implementation. The layer has dimension $(OC,IC,K,K) = (256,128,3,3)$, and the input's dimensions are set to $(B,IC,M,N) = (8,128,7,7)$ in (a) and (b) and to $(B,IC,M,N) = (32,256,244,244)$ in (c) and (d).}%
\label{fig:conv-par-run}
\end{center}
\end{figure}

This section analyzes the runtime scaling for our linear and convolutional layers implementations of Algorithms~\ref{alg:lin-backprop} and~\ref{alg:conv-backprop}. 
To validate our linear-scaling claims, we specifically examine the runtime dependence of our implementations relative to sparsity. 
These implementations are compared to their dense counterparts available in Pytorch. Additionally, for the linear layer implementation, we compare it against the sparse implementation offered by Pytorch. We present the improvement in our implementations' performance as a function of a layer's sparsity ranging from $80\%$ to $99\%$. In particular, we give the runtime for our multithreaded implementations run on $8$-threads in Figures~\ref{fig:conv-par-run} and~\ref{fig:lin-par-run}, and the runtime for the single-core implementations in Figure~\ref{fig:single-run}. To highlight the speedup between implementations, we give all our measurements in \emph{log-lin} plots.

\paragraph{Linear layers.}
We evaluate the performance of our linear layer by measuring the runtime of forward and backward pass through a large layer of dimensions $(M,N) = (768,3072)$ and an input of size $(B,M) = (902, 768)$. We report the single-core results for the backward pass in Figure~\ref{subfig:lin-single-run}, and we compare them to a Pytorch-based sparse implementation, which is currently only single-core on CPUs. Our sequential performance increases linearly with the sparsity of $\mathbf{W}$, we match a dense implementation around $90\%$ sparsity, and we obtain a $5\times$ speedup at $99\%$. In Figure~\ref{fig:conv-par-run}, we plot the results for the multithreaded forward and backward passes. For our multithreaded forward pass, we observe similar behavior to the single-core variant beating the dense implementation after $95\%$ sparsity.
It should be noted that the performance of our multithreaded implementation of the backward pass is currently  limited by synchronization overheads, which can be removed with additional optimizations. As of now, it matches the dense implementation only at very high levels of sparsity.

\paragraph{Convolutional layers.}
In the case of the convolutional layers, we present the runtime performance for two distinct scenarios: small input tensor and large input tensor. The dimensions of the layer are set to $(OC,IC,K,K) = (256,128,3,3)$, and the input tensor's dimensions are set to $(B,IC,M,N) = (8,128,7,7)$ and $(B,IC,M,N) = (32,256,244,244)$. These scenarios highlight the difference between the two sparse convolution kernels we developed: $\texttt{SparseConv}$ and $\texttt{SparseConvOverON}$. The former is presented in Algorithm~\ref{alg:conv-backprop}, and the latter is its permuted variant that vectorizes over the $ON$ dimension of the tensor.
 
We show that by permuting the input, we substantially improve the performance of our algorithm. The single-core runtime performance over a small input is presented in Figure~\ref{subfig:conv-single-run}, where we observe a significant speedup compared to a dense implementation, with $19$x speedup at $99\%$ sparsity.
The parallel runtime for forward and backward passes are in Figure~\ref{fig:conv-par-run}. Timings for small input sizes are given in Figures~\ref{subfig:par-a} and~\ref{subfig:par-b}, and for large inputs in Figures~\ref{subfig:par-c} and~\ref{subfig:par-d}. We see how permuting impacts our performance for small $ON$ values achieving a speedup over the dense implementation of up to $5$x for the forward and backward pass at $99\%$ sparsity. On the other hand, for larger $M$ and $N$, our results indicate that vectorizing over $ON$ yields the best performance, resulting in a speedup of $3.35$x for the forward pass and $9$x for the backward.

\subsection{End-to-End Training Experiments}

We now evaluate SparseProp on \emph{sparse transfer learning}, and \emph{sparse training from scratch}. In each case we examine sparsity settings which, while maintaining reasonable accuracy, can achieve non-trivial speedups. 
Since we executed over more than 15 different tasks, on multiple models, the full experiments used to determine accuracy are executed on GPU. At the same time, we computed CPU speedups using proportionally-shortened versions of the training schedules on CPU, and validated correctness in a few end-to-end runs on different tasks. 

In all the experiments, dense modules (linear or convolution) are executed dense as long as they are less than $80\%$ sparse. Once a module reaches at least $80\%$ sparsity (which may happen during training for gradual pruning scenarios), we measure the time to run one batch through both dense and sparse versions, and choose the fastest version based on forward+backward time (in the case of convolution, both sparse implementations are considered). 
Notice that in all experiments, the sparsity patterns change only a few times during the whole run, meaning the overhead of these few extra batches is negligible. 
(An alternative approach would be to generate a static database of the best implementation choices for each layer type and size, and greedily adopt the implementation for each layer in turn.)

\subsubsection{Application 1: Sparse Transfer}

\paragraph{Image Classification.} We first consider a standard transfer learning setup for image classification using CNNs, in which a model pretrained on ImageNet-1K has its last (FC) layer resized and re-initialized, and then is further finetuned on a smaller target dataset. 
As targets, we focus on twelve datasets that are conventionally used as benchmarks for transfer learning, e.g. in~\cite{kornblith2019better, salman2020adversarially, iofinova2022well}. See Table \ref{table:datasets} for a summary of the tasks. Importantly, input images are scaled to standard ImageNet size, i.e. $224 \times 224 \times 3$, resulting in proportional computational costs. 

\begin{figure}
    \centering
    \includegraphics[width=0.42\textwidth]{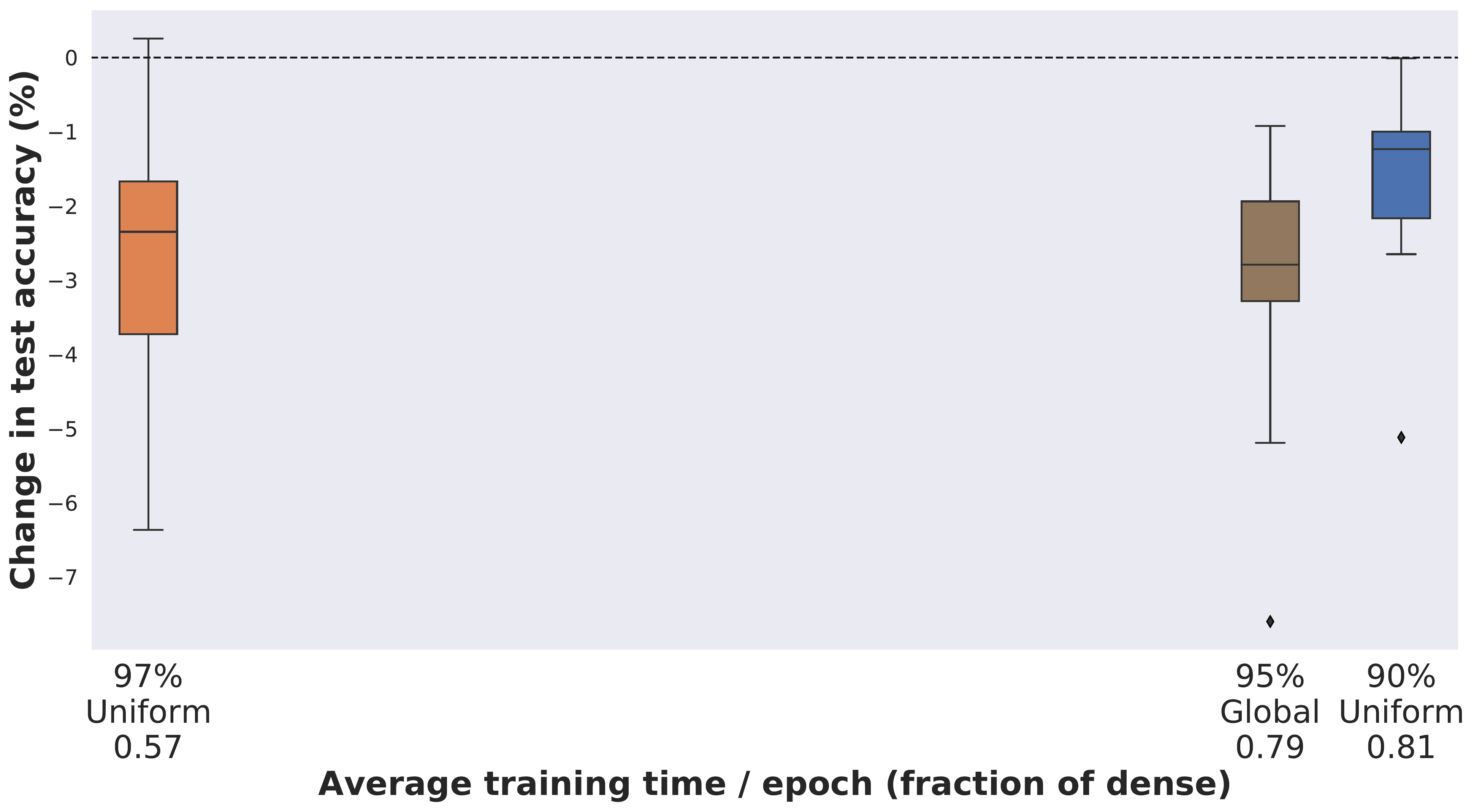}
    \caption{Accuracy vs. speedup for \emph{transfer learning} experiments on the ResNet50 architecture. The boxplots represent aggregated performance across all twelve target tasks.}
    \label{fig:img_time_vs_acc_transfer}
\end{figure}

\begin{figure}
    \centering
    \includegraphics[width=0.42\textwidth]{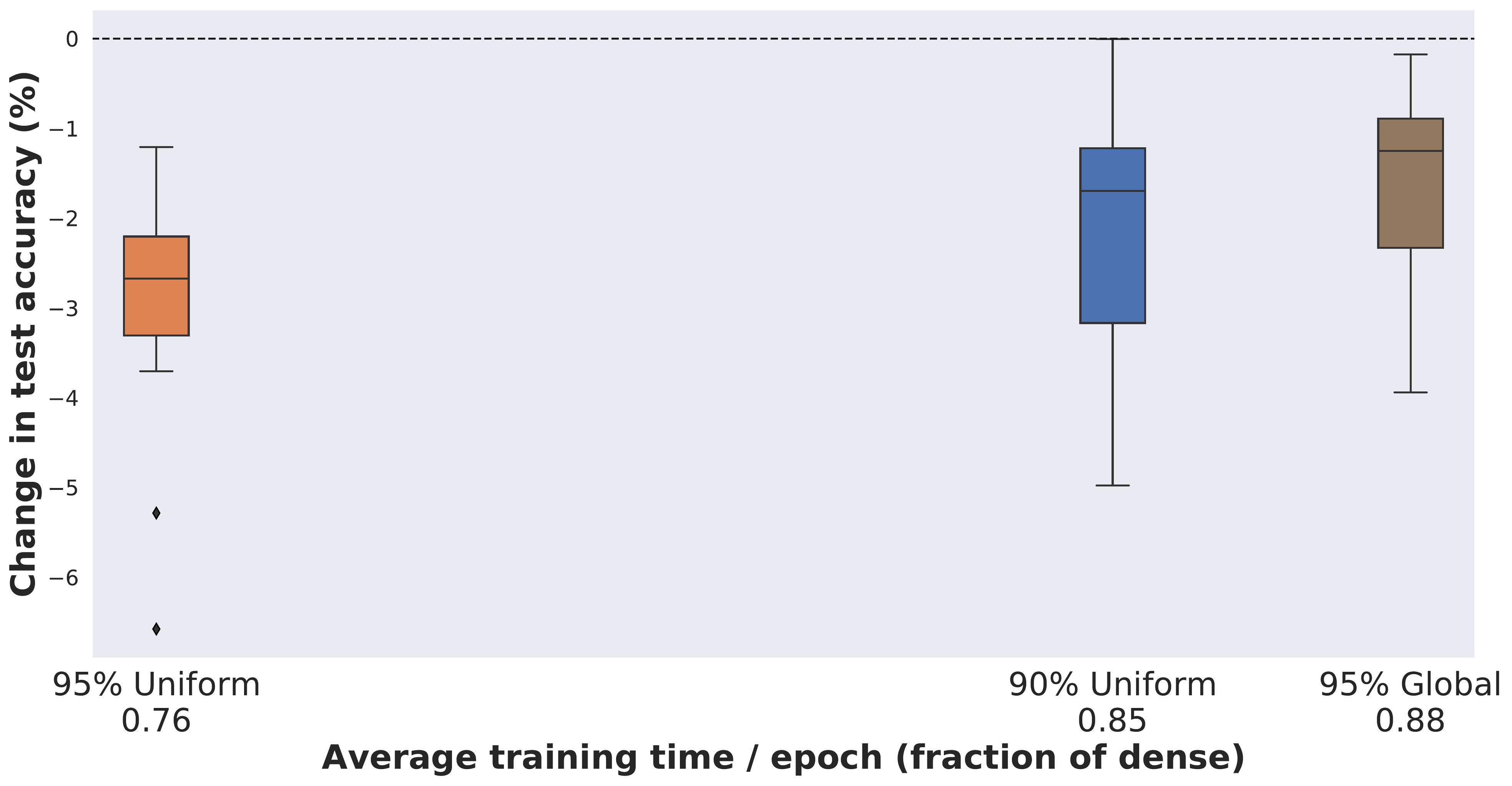}
    \caption{Accuracy vs. estimated speedup for \emph{from-scratch learning} experiments on the ResNet18 architecture. The boxplots represent aggregated performance across all twelve target tasks.}
    \label{fig:img_time_vs_acc_scratch}
\end{figure}

We consider dense and sparse ResNet50 models pre-trained (and sparsified) on the ImageNet-1K dataset. 
Models are pruned using the AC/DC method~\cite{peste2021ac}, which shows high accuracy on ImageNet-1K, and produces transferrable sparse features~\cite{iofinova2022well}. (We adopt their publicly-available models.) 
To explore the accuracy-vs-speedup trade-off, we consider both a \emph{Uniform pruning} scenario, in which all convolutional layers except for the input are pruned to a uniform sparsity (90\% and 97\%), and a \emph{Global pruning} scenario, in which all convolutional layers are pruned jointly, to a target average sparsity (95\%), using the global magnitude pruning criterion. The former two models have been trained for 200 epochs each, and have 76.01\% and 74.12\% Top-1 accuracy, whereas the latter is trained for 100 epochs and has 73.1\% Top-1 accuracy. 
The dense model has 76.8\% Top-1 accuracy. 
For transfer learning, we maintain the sparsity pattern of the models, reinitialize the final fully-connected layer, and train for 150 epochs on each of the 12 ``downstream'' tasks.

In Figure~\ref{fig:img_time_vs_acc_transfer}, we aggregated results across all tasks, in terms of mean and variance of the \emph{accuracy drop relative to transferring the dense model}
(the full per-task accuracy results are presented in Table~\ref{table:rn50_full_all}, and speedups are presented in Table~\ref{tab:rn50-finetune-speedups}). 
As expected, the aggregated sparse test accuracy drops relative to the dense baseline, 
proportionally to the ImageNet Top-1 accuracy. 
The Uniform-90 model shows the smallest drops (1\% on average), but also the lowest end-to-end speedup (25\%), while the Uniform-97 and Global-95 models have slightly worse average drops (around 2\%). 
Remarkably, due to higher initial accuracy, the Uniform-97 model has similar accuracy to Global-95, 
but much higher end-to-end speedup of 1.75x.

\begin{figure}[t]
    \centering
    \includegraphics[width=0.45\textwidth]{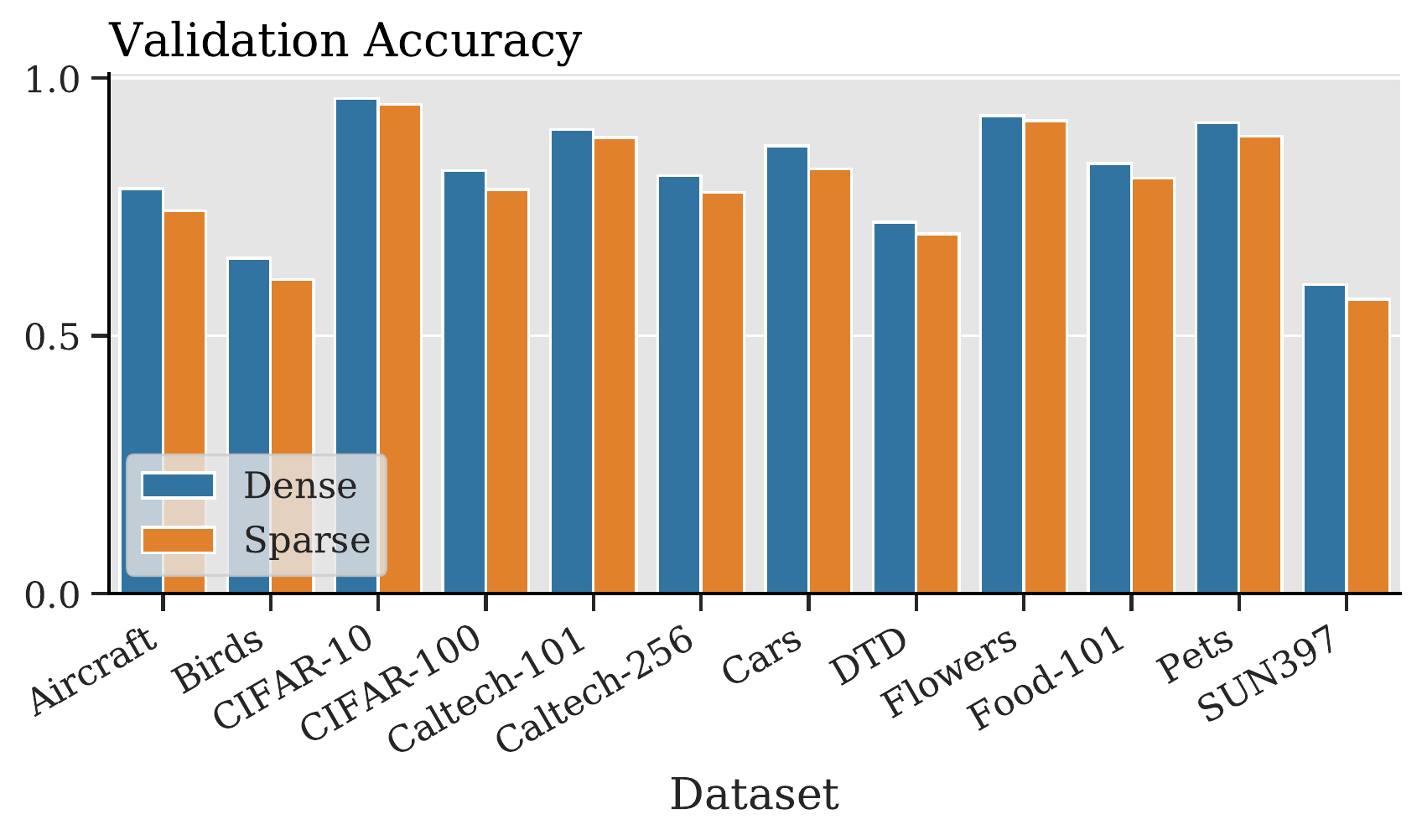}
    \caption{Top-1 validation transfer accuracy of dense and 95\% sparse ResNet18 models pre-trained on ImageNet-1K.}
    \label{fig:rn18_transfer_dset_results}
\end{figure}

\paragraph{Case Study: 95\% Uniformly-Pruned ResNet18.} 
We now analyze in detail both the accuracy drops and the per-layer and global speedups for a 95\% uniformly-pruned ResNet18 model. 
On ImageNet, the AC/DC pruned model ResNet18 model has 68.9\% Top-1 accuracy, a relative 1\% drop from the Torchvision baseline. 
Figure~\ref{fig:rn18_transfer_dset_results} depicts the transfer performance of the respective models on twelve target datasets, using exactly the same transfer recipe for both sparse and dense models. 
The accuracy loss is moderate (2.85\% average, with a maximum gap of 4.5\% Top-1, across all tasks). 

\begin{figure}[t]
    \centering
    \includegraphics[width=0.45\textwidth]{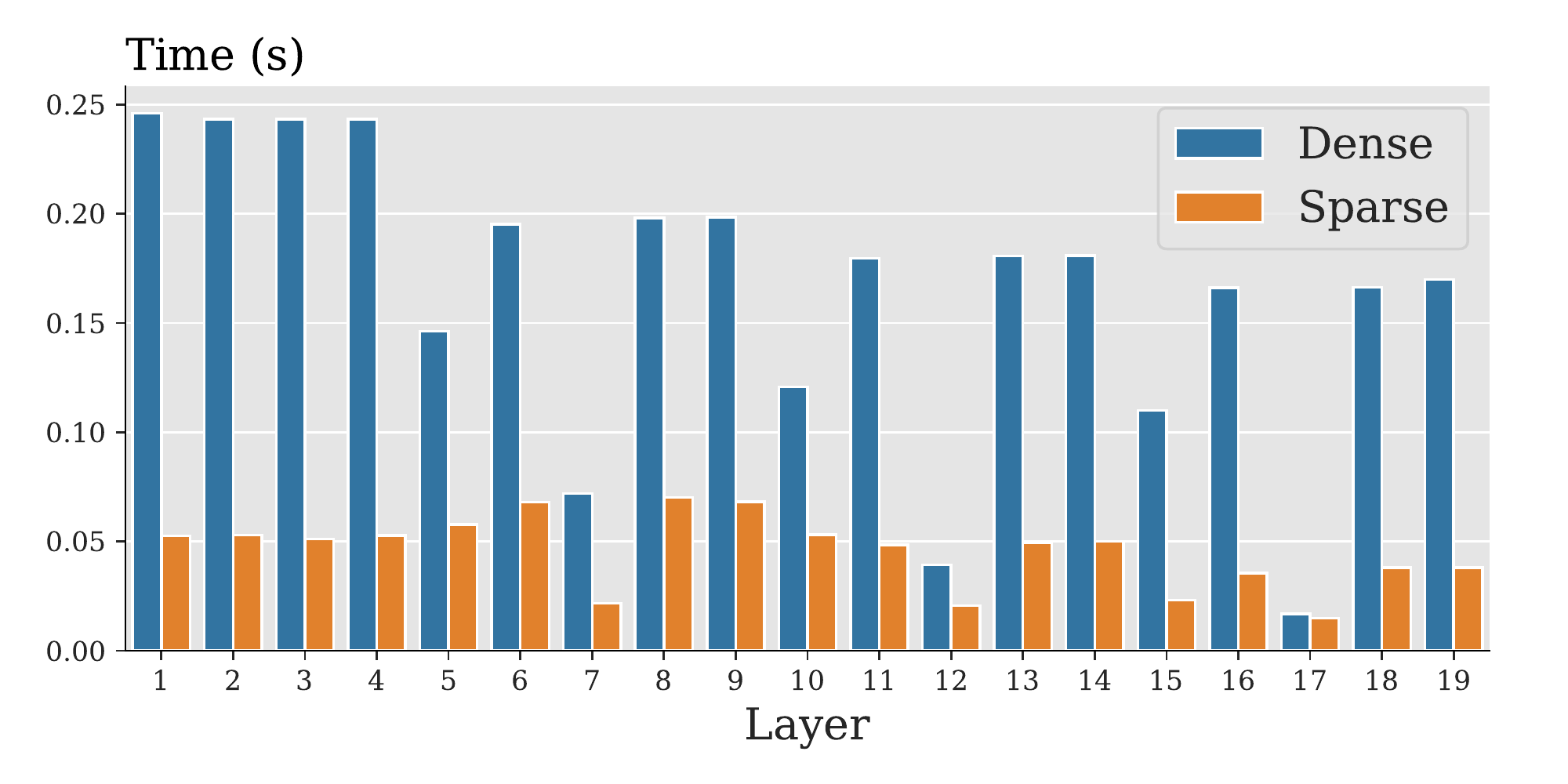}
    \caption{Layer-wise back-propagation time comparison between Dense and 95\% uniformly-pruned Resnet18. Note that first and last layers are always dense and are hence removed from the comparison.}
    \label{fig:rn18_95_un_back}
\end{figure}

Figure~\ref{fig:rn18_95_un_back} depicts layer-wise backward speedups with respect to Pytorch's dense implementation. 
The results show that the overall end-to-end speedup is 1.58x, with a 1.26x speedup end-to-end for forward computations and a 2.11x speedup end-to-end for backward computations. A closer examination reveals that if we only measure the time spent in convolution and linear modules' forward and backward functions we get a 2.53x speedup, suggesting the presence of significant overheads outside of the convolution and linear computations (such as batch normalization layers and ReLU) for the Pytorch CPU implementation. More precisely, our implementations provide speedups of 1.57x and 3.60x, for the forward and backward multiplications, respectively. This highlights the  efficiency of our backward algorithms, but also the overheads of 
Pytorch's current CPU training implementation. (Specifically, in our sparse implementation, the batch normalization and ReLU computations, executed via Pytorch, take approximately 25\% of the total training time.)

\paragraph{Language Modelling.} Next, we replicate the setup of~\citet{kurtic2022optimal}, where a BERT-base~\citep{devlin2018bert} model is pruned in the pre-training stage on BookCorpus and English Wikipedia~\citep{hf-datasets} with the state-of-the-art unstructured pruner oBERT~\citep{kurtic2022optimal}. After that, the remaining weights are fine-tuned on several downstream tasks with fixed sparsity masks. We consider a model with 97\% global sparsity, and maintain its masks throughout finetuning on the MNLI and QQP tasks from the GLUE benchmark~\cite{wang2018glue}.
Both accuracy and speedup results are shown in Table~\ref{tab:bert-transfer}, and show 37\% speedup on a single core for inference on this model, at the price of $\sim$ 1--3.5\% accuracy drop. 

\begin{table}\centering
\ra{1.3}
\scalebox{0.42}{
\begin{tabular}{@{}c|ccc|cccc@{}}\toprule
ResNet50, batch size=$64$ 
& Forward & Backward & End-to-End && Forward & Backward & End-to-End\\ \midrule
Dense &$10.98s \pm 0.14s$&$19.85s \pm 0.10s$&$\mathbf{31.19s \pm 0.24s}$&&$1.44s \pm 0.09s$&$3.41s \pm 0.04s$&$\mathbf{5.09s \pm 0.09s}$\\
Uniform $90\% $&$1.11\times$&$1.31\times$&$\mathbf{1.23\times}$&&  $1.00\times$&$1.26\times$&$\mathbf{1.16\times}$\\
Global $95\%  $&$1.24\times$&$1.28\times$&$\mathbf{1.27\times}$&&  $1.13\times$&$1.18\times$&$\mathbf{1.15\times}$\\
Uniform $97\% $&$1.59\times$&$1.88\times$&$\mathbf{1.76\times}$&&  $1.40\times$&$1.76\times$&$\mathbf{1.59\times}$\\
Global $98\%  $&$1.42\times$&$1.71\times$&$\mathbf{1.58\times}$&&  $1.29\times$&$1.61\times$&$\mathbf{1.46\times}$\\
\bottomrule
\end{tabular}
}
\caption{Single-core (left) and parallel (right) relative speedup for transfer learning on sparse ResNet50 models pretrained on ImageNet1k over a dense implementation.}
\label{tab:rn50-finetune-speedups}
\end{table}

\begin{table}\centering
\ra{1.3}
\scalebox{0.60}{
\begin{tabular}{@{}c|cc|c|cc@{}}\toprule
BERT-base 
& Forward & Backward & End-to-End & MNLI & QQP\\ \midrule
Dense &$0.72s \pm 0.01s$&$1.33s \pm 0.00s$&$\mathbf{2.67s \pm 0.01s}$&$84.54\%$&$91.06\%$\\
Global $97\%  $&$1.27\times$&$1.40\times$&$\mathbf{1.37\times}$&$80.91\%$& $90.33\%$\\
\bottomrule
\end{tabular}
}
\caption{Accuracies and single-core relative speedup for transfer learning sparse BERT-base models.}
\label{tab:bert-transfer}
\end{table}

\label{subsection:transfer}

\subsubsection{Application 2: Sparse Training}

\begin{table}\centering

\newcommand{\miss}{-}
\ra{1.3}
\scalebox{0.50}{
\begin{tabular}{@{}ccccccccccc@{}}\toprule
&\multicolumn{4}{c}{ResNet18, batch size=$256$}&\multicolumn{4}{c}{BERT-base, batch size=$4$}\\
\cmidrule{2-5} \cmidrule{6-9}
& Forward & Backward & End-to-End&CelebA AUC&Forward & Backward & End-to-End & QQP\\ \midrule
Dense         &$1\times$&$1\times$&$1\times$ & 80.2 &$1\times$ &$1\times$ &$1\times$ &91.06\\
 Global $90\%  $&$1.02\times$&$1.14\times$&$\mathbf{1.08\times}$&81.6&\miss&\miss&\miss&\miss\\
Uniform $90\% $&$0.99\times$&$1.36\times$&$\mathbf{1.17\times}$&\miss&$0.97\times$&$0.96\times$&$\mathbf{1.04\times}$&$90.09$\\
Global $95\%  $&$1.04\times$&$1.23\times$&$\mathbf{1.13\times}$&81.7&\miss&\miss&\miss&\miss\\
Uniform $95\% $&$1.09\times$&$1.57\times$&$\mathbf{1.32\times}$&\miss&$1.16\times$&$1.19\times$&$\mathbf{1.23\times}$&$89.15$\\
Uniform $97\% $&$1.19\times$&$1.65\times$&$\mathbf{1.41\times}$&81.1&$1.24\times$&$1.31\times$&$\mathbf{1.31\times}$&$87.81$\\
Global $99\%  $&$1.08\times$&$1.48\times$&$\mathbf{1.28\times}$&81.0&\miss&\miss&\miss&\miss\\
\bottomrule
\end{tabular}
}
\caption{Relative speedup on the from-scratch training for sparse ResNet18 and BERT-base models over a dense implementation.}
\label{tab:speedup_table}
\end{table}

\paragraph{Image Classification.} 
Finally, we evaluate SparseProp in the from-scratch sparse training scenario. 
We first consider the same 12 specialized datasets as in Section~\ref{subsection:transfer}, 
and train a ResNet18 architecture \emph{from scratch}. 
We apply 90\% and 95\% sparsity using Gradual Magnitude Pruning \cite{zhu2017prune}, 
in two scenarios---Uniform sparsity, in which all convolutional layers, except the first, are pruned to the same target, and the Global scenario, in which the magnitudes of the weights are aggregated across all layers, and the smallest-magnitude weights are pruned to reach the target sparsity. (For the latter scenario, we consider only 95\% sparse models.) Note that in the Uniform scenario, we do not prune the initial convolution, nor the final FC layer, while in the Global scenario, we do. In both cases, we train the model dense for ten epochs before pruning the lowest 5\% of the weights (either uniformly or globally) and then prune gradually every 10 epochs until epoch 80, at which point we fine-tune for a further 20 epochs.

The results are given in Figure~\ref{fig:img_time_vs_acc_scratch}. In terms of accuracy, Global-GMP outperforms Uniform-GMP at the same target sparsity, with Global-95\% showing similar accuracy to Uniform-90\%, though in all cases performance is inferior to when ImageNet weights are used for pre-training. The highest speedup is of 1.31x,  for 95\% Uniform sparsity.  

Additionally, we evaluate SparseProp on the CelebA dataset~\cite{Liu2015DeepLF}, which consists of a training set of 162'770 images, and a validation set of 19'962 images from 10 000 well-known individuals, each annotated with forty binary attributes, such as "Smiling", "Male", "Wearing Necklace", etc. 
We consider the task of jointly predicting all forty attributes, in a similar setup as above, and show the resulting AUC in Table~\ref{tab:speedup_table}. We observe that AUC stays fairly constant even at high sparsities, even as the speed of training increases.

\paragraph{Language Modelling.} For sparse fine-tuning from scratch on language models, we start from the pre-trained BERT-base~\citep{devlin2018bert} model, which we fine-tune for 3 epochs on the target downstream task, and then prune in \textit{one-shot} with the state-of-the-art unstructured pruner oBERT~\citep{kurtic2022optimal}, uniformly to 90\%, 95\% or 97\% per-layer sparsity.  
After one-shot pruning, we fine-tune the remaining weights for 5 epochs and examine accuracy and speedup versus the dense variant. The results are presented in Table~\ref{tab:speedup_table} (right), and show  end-to-end speedups of up to 30\%, at an accuracy loss between 1 and 3.5\%. 

\label{subsection:scratch}

% \vspace{-1em}
\section{Discussion}

We have provided an efficient vectorized algorithm for sparse backpropagation, 
with linear runtime dependency in the density of the layer weights. 
We have also provided an efficient CPU-based implementation of this algorithm, 
and integrated it with the popular Pytorch framework. 
Experimental evidence validates the runtime scaling of our algorithm on various layer shapes and types. 
We complemented this algorithmic contribution with an extensive study of the feasibility of 
sparse transfer learning and from-scratch training in edge scenarios. 
We observed consistent speedups across scenarios, at the cost of moderate accuracy loss. 
Our results should serve as motivation for further research into accurate sparse training in this setting, 
in particular for leveraging sparsity on highly-specialized tasks, which is an under-studied area.  

\bibliography{references.bib}

\begin{thebibliography}{64}
\providecommand{\natexlab}[1]{#1}
\providecommand{\url}[1]{\texttt{#1}}
\expandafter\ifx\csname urlstyle\endcsname\relax
  \providecommand{\doi}[1]{doi: #1}\else
  \providecommand{\doi}{doi: \begingroup \urlstyle{rm}\Url}\fi

\bibitem[Bellec et~al.(2018)Bellec, Kappel, Maass, and
  Legenstein]{bellec2017deep}
Bellec, G., Kappel, D., Maass, W., and Legenstein, R.
\newblock Deep rewiring: Training very sparse deep networks.
\newblock \emph{International Conference on Learning Representations (ICLR)},
  2018.

\bibitem[Berg et~al.(2014)Berg, Liu, Lee, Alexander, Jacobs, and
  Belhumeur]{berg2014birds}
Berg, T., Liu, J., Lee, S.~W., Alexander, M.~L., Jacobs, D.~W., and Belhumeur,
  P.~N.
\newblock Birdsnap: Large-scale fine-grained visual categorization of birds.
\newblock In \emph{Conference on Computer Vision and Pattern Recognition
  (CVPR)}, 2014.

\bibitem[Bossard et~al.(2014)Bossard, Guillaumin, and
  Van~Gool]{bossard2014food101}
Bossard, L., Guillaumin, M., and Van~Gool, L.
\newblock Food-101 -- mining discriminative components with random forests.
\newblock In \emph{European Conference on Computer Vision (ECCV)}, 2014.

\bibitem[Chen et~al.(2021)Chen, Frankle, Chang, Liu, Zhang, Carbin, and
  Wang]{chen2021lottery}
Chen, T., Frankle, J., Chang, S., Liu, S., Zhang, Y., Carbin, M., and Wang, Z.
\newblock The lottery tickets hypothesis for supervised and self-supervised
  pre-training in computer vision models.
\newblock In \emph{Proceedings of the IEEE/CVF Conference on Computer Vision
  and Pattern Recognition}, pp.\  16306--16316, 2021.

\bibitem[Cimpoi et~al.(2014)Cimpoi, Maji, Kokkinos, Mohamed, and
  Vedaldi]{cimpoi2014dtd}
Cimpoi, M., Maji, S., Kokkinos, I., Mohamed, S., and Vedaldi, A.
\newblock Describing textures in the wild.
\newblock In \emph{Conference on Computer Vision and Pattern Recognition
  (CVPR)}, 2014.

\bibitem[Dettmers \& Zettlemoyer(2019)Dettmers and
  Zettlemoyer]{dettmers2019sparse}
Dettmers, T. and Zettlemoyer, L.
\newblock Sparse networks from scratch: Faster training without losing
  performance.
\newblock \emph{arXiv preprint arXiv:1907.04840}, 2019.

\bibitem[Devlin et~al.(2019)Devlin, Chang, Lee, and Toutanova]{devlin2018bert}
Devlin, J., Chang, M.-W., Lee, K., and Toutanova, K.
\newblock {BERT}: Pre-training of deep bidirectional transformers for language
  understanding.
\newblock In \emph{North American Chapter of the Association for Computational
  Linguistics (NAACL)}, 2019.

\bibitem[Elsen et~al.(2020)Elsen, Dukhan, Gale, and Simonyan]{elsen2020fast}
Elsen, E., Dukhan, M., Gale, T., and Simonyan, K.
\newblock Fast sparse convnets.
\newblock In \emph{Conference on Computer Vision and Pattern Recognition
  (CVPR)}, 2020.

\bibitem[Evci et~al.(2020)Evci, Gale, Menick, Castro, and
  Elsen]{evci2020rigging}
Evci, U., Gale, T., Menick, J., Castro, P.~S., and Elsen, E.
\newblock Rigging the lottery: Making all tickets winners.
\newblock In \emph{International Conference on Machine Learning (ICML)}, 2020.

\bibitem[Gale et~al.(2020)Gale, Zaharia, Young, and Elsen]{gale2020sparse}
Gale, T., Zaharia, M., Young, C., and Elsen, E.
\newblock Sparse gpu kernels for deep learning.
\newblock In \emph{SC20: International Conference for High Performance
  Computing, Networking, Storage and Analysis}, pp.\  1--14. IEEE, 2020.

\bibitem[Gray et~al.(2017)Gray, Radford, and Kingma]{gray2017gpu}
Gray, S., Radford, A., and Kingma, D.~P.
\newblock Gpu kernels for block-sparse weights.
\newblock \emph{arXiv preprint arXiv:1711.09224}, 3:\penalty0 2, 2017.

\bibitem[Griffin et~al.(2006)Griffin, Holub, and Perona]{holub2006caltech256}
Griffin, G., Holub, A.~D., and Perona, P.
\newblock {The Caltech 256}.
\newblock \emph{Caltech Technical Report}, 2006.

\bibitem[Hagiwara(1994)]{hagiwara1994}
Hagiwara, M.
\newblock A simple and effective method for removal of hidden units and
  weights.
\newblock \emph{Neurocomputing}, 6\penalty0 (2):\penalty0 207 -- 218, 1994.
\newblock ISSN 0925-2312.
\newblock Backpropagation, Part IV.

\bibitem[Han et~al.(2016{\natexlab{a}})Han, Liu, Mao, Pu, Pedram, Horowitz, and
  Dally]{han2016eie}
Han, S., Liu, X., Mao, H., Pu, J., Pedram, A., Horowitz, M.~A., and Dally,
  W.~J.
\newblock Eie: Efficient inference engine on compressed deep neural network.
\newblock \emph{ACM SIGARCH Computer Architecture News}, 44\penalty0
  (3):\penalty0 243--254, 2016{\natexlab{a}}.

\bibitem[Han et~al.(2016{\natexlab{b}})Han, Mao, and Dally]{han2015deep}
Han, S., Mao, H., and Dally, W.~J.
\newblock Deep compression: Compressing deep neural networks with pruning,
  trained quantization and {Huffman} coding.
\newblock In \emph{International Conference on Learning Representations
  (ICLR)}, 2016{\natexlab{b}}.

\bibitem[He et~al.(2016)He, Zhang, Ren, and Sun]{he2016deep}
He, K., Zhang, X., Ren, S., and Sun, J.
\newblock Deep residual learning for image recognition.
\newblock In \emph{Conference on Computer Vision and Pattern Recognition
  (CVPR)}, 2016.

\bibitem[Hoefler et~al.(2021)Hoefler, Alistarh, Ben-Nun, Dryden, and
  Peste]{hoefler2021sparsity}
Hoefler, T., Alistarh, D., Ben-Nun, T., Dryden, N., and Peste, A.
\newblock Sparsity in deep learning: Pruning and growth for efficient inference
  and training in neural networks.
\newblock \emph{arXiv preprint arXiv:2102.00554}, 2021.

\bibitem[Hubara et~al.(2021)Hubara, Chmiel, Island, Banner, Naor, and
  Soudry]{hubara2021accelerated}
Hubara, I., Chmiel, B., Island, M., Banner, R., Naor, S., and Soudry, D.
\newblock Accelerated sparse neural training: A provable and efficient method
  to find {N:M} transposable masks.
\newblock In \emph{Conference on Neural Information Processing Systems
  (NeurIPS)}, 2021.

\bibitem[Iofinova et~al.(2022)Iofinova, Peste, Kurtz, and
  Alistarh]{iofinova2022well}
Iofinova, E., Peste, A., Kurtz, M., and Alistarh, D.
\newblock How well do sparse {ImageNet} models transfer?
\newblock In \emph{Conference on Computer Vision and Pattern Recognition
  (CVPR)}, 2022.

\bibitem[Ivanov et~al.(2022)Ivanov, Dryden, and Hoefler]{ivanov2022sten}
Ivanov, A., Dryden, N., and Hoefler, T.
\newblock Sten: An interface for efficient sparsity in pytorch.
\newblock 2022.

\bibitem[Jakob et~al.(2016)Jakob, Rhinelander, and Moldovan]{pybind11}
Jakob, W., Rhinelander, J., and Moldovan, D.
\newblock pybind11 — seamless operability between c++11 and python, 2016.
\newblock URL \url{https://github.com/pybind/pybind11}.

\bibitem[Jayakumar et~al.(2020)Jayakumar, Pascanu, Rae, Osindero, and
  Elsen]{jayakumar2020top}
Jayakumar, S., Pascanu, R., Rae, J., Osindero, S., and Elsen, E.
\newblock {Top-KAST}: {Top-K} always sparse training.
\newblock In \emph{Conference on Neural Information Processing Systems
  (NeurIPS)}, 2020.

\bibitem[Jiang et~al.(2022)Jiang, Hu, and Song]{jiangexposing}
Jiang, P., Hu, L., and Song, S.
\newblock Exposing and exploiting fine-grained block structures for fast and
  accurate sparse training.
\newblock In \emph{Advances in Neural Information Processing Systems}, 2022.

\bibitem[Kornblith et~al.(2019)Kornblith, Shlens, and Le]{kornblith2019better}
Kornblith, S., Shlens, J., and Le, Q.~V.
\newblock Do better imagenet models transfer better?
\newblock In \emph{Proceedings of the IEEE/CVF conference on computer vision
  and pattern recognition}, pp.\  2661--2671, 2019.

\bibitem[Krause et~al.(2013)Krause, Stark, Deng, and Fei-Fei]{krause2013cars}
Krause, J., Stark, M., Deng, J., and Fei-Fei, L.
\newblock {3D Object Representations for Fine-Grained Categorization}.
\newblock In \emph{4th International IEEE Workshop on 3D Representation and
  Recognition}, Sydney, Australia, 2013.

\bibitem[Krizhevsky et~al.(2009)Krizhevsky, Hinton, et~al.]{cifar100}
Krizhevsky, A., Hinton, G., et~al.
\newblock Learning multiple layers of features from tiny images.
\newblock 2009.

\bibitem[Kurtic et~al.(2022)Kurtic, Campos, Nguyen, Frantar, Kurtz, Fineran,
  Goin, and Alistarh]{kurtic2022optimal}
Kurtic, E., Campos, D., Nguyen, T., Frantar, E., Kurtz, M., Fineran, B., Goin,
  M., and Alistarh, D.
\newblock The optimal bert surgeon: Scalable and accurate second-order pruning
  for large language models.
\newblock In \emph{Proceedings of the 2022 Conference on Empirical Methods in
  Natural Language Processing (EMNLP)}, pp.\  4163--–4181, 2022.

\bibitem[Kusupati et~al.(2020)Kusupati, Ramanujan, Somani, Wortsman, Jain,
  Kakade, and Farhadi]{kusupati2020soft}
Kusupati, A., Ramanujan, V., Somani, R., Wortsman, M., Jain, P., Kakade, S.,
  and Farhadi, A.
\newblock Soft threshold weight reparameterization for learnable sparsity.
\newblock In \emph{International Conference on Machine Learning (ICML)}, 2020.

\bibitem[Lagunas et~al.(2021)Lagunas, Charlaix, Sanh, and
  Rush]{lagunas2021block}
Lagunas, F., Charlaix, E., Sanh, V., and Rush, A.~M.
\newblock Block pruning for faster transformers.
\newblock In \emph{Conference on Empirical Methods in Natural Language
  Processing (EMNLP)}, 2021.

\bibitem[LeCun et~al.(1990)LeCun, Denker, and Solla]{lecun1990optimal}
LeCun, Y., Denker, J.~S., and Solla, S.~A.
\newblock Optimal brain damage.
\newblock In \emph{Conference on Neural Information Processing Systems
  (NeurIPS)}, 1990.

\bibitem[Lhoest et~al.(2021)Lhoest, Villanova~del Moral, Jernite, Thakur, von
  Platen, Patil, Chaumond, Drame, Plu, Tunstall, Davison, {\v{S}}a{\v{s}}ko,
  Chhablani, Malik, Brandeis, Le~Scao, Sanh, Xu, Patry, McMillan-Major, Schmid,
  Gugger, Delangue, Matussi{\`e}re, Debut, Bekman, Cistac, Goehringer, Mustar,
  Lagunas, Rush, and Wolf]{hf-datasets}
Lhoest, Q., Villanova~del Moral, A., Jernite, Y., Thakur, A., von Platen, P.,
  Patil, S., Chaumond, J., Drame, M., Plu, J., Tunstall, L., Davison, J.,
  {\v{S}}a{\v{s}}ko, M., Chhablani, G., Malik, B., Brandeis, S., Le~Scao, T.,
  Sanh, V., Xu, C., Patry, N., McMillan-Major, A., Schmid, P., Gugger, S.,
  Delangue, C., Matussi{\`e}re, T., Debut, L., Bekman, S., Cistac, P.,
  Goehringer, T., Mustar, V., Lagunas, F., Rush, A., and Wolf, T.
\newblock Datasets: A community library for natural language processing.
\newblock In \emph{Proceedings of the 2021 Conference on Empirical Methods in
  Natural Language Processing: System Demonstrations}, pp.\  175--184.
  Association for Computational Linguistics, November 2021.

\bibitem[Li et~al.(2004)Li, Fergus, and Perona]{li2004caltech101}
Li, F.-F., Fergus, R., and Perona, P.
\newblock Learning generative visual models from few training examples: an
  incremental {Bayesian} approach tested on 101 object categories.
\newblock In \emph{Conference on Computer Vision and Pattern Recognition
  (CVPR)}, 2004.

\bibitem[Lis et~al.(2019)Lis, Golub, and Lemieux]{lis2019full}
Lis, M., Golub, M., and Lemieux, G.
\newblock Full deep neural network training on a pruned weight budget.
\newblock \emph{Proceedings of Machine Learning and Systems}, 1:\penalty0
  252--263, 2019.

\bibitem[Liu et~al.(2015)Liu, Luo, Wang, and Tang]{Liu2015DeepLF}
Liu, Z., Luo, P., Wang, X., and Tang, X.
\newblock Deep learning face attributes in the wild.
\newblock \emph{2015 IEEE International Conference on Computer Vision (ICCV)},
  2015.

\bibitem[Maji et~al.(2013)Maji, Rahtu, Kannala, Blaschko, and
  Vedaldi]{maji13fgvc-aircraft}
Maji, S., Rahtu, E., Kannala, J., Blaschko, M., and Vedaldi, A.
\newblock Fine-grained visual classification of aircraft.
\newblock \emph{arXiv preprint arXiv:1306.5151}, 2013.

\bibitem[Merity et~al.(2016)Merity, Xiong, Bradbury, and Socher]{wikitext103}
Merity, S., Xiong, C., Bradbury, J., and Socher, R.
\newblock Pointer sentinel mixture models.
\newblock \emph{arXiv preprint arXiv:1609.07843}, 2016.

\bibitem[Mishra et~al.(2021)Mishra, Latorre, Pool, Stosic, Stosic, Venkatesh,
  Yu, and Micikevicius]{NVIDIASparse}
Mishra, A., Latorre, J.~A., Pool, J., Stosic, D., Stosic, D., Venkatesh, G.,
  Yu, C., and Micikevicius, P.
\newblock Accelerating sparse deep neural networks.
\newblock \emph{arXiv preprint arXiv:2104.08378}, 2021.

\bibitem[Mocanu et~al.(2016)Mocanu, Mocanu, Nguyen, Gibescu, and
  Liotta]{mocanu2016topological}
Mocanu, D.~C., Mocanu, E., Nguyen, P.~H., Gibescu, M., and Liotta, A.
\newblock A topological insight into restricted boltzmann machines.
\newblock \emph{Machine Learning}, 104:\penalty0 243--270, 2016.

\bibitem[Mostafa \& Wang(2019)Mostafa and Wang]{mostafa2019parameter}
Mostafa, H. and Wang, X.
\newblock Parameter efficient training of deep convolutional neural networks by
  dynamic sparse reparameterization.
\newblock In \emph{International Conference on Machine Learning}, pp.\
  4646--4655. PMLR, 2019.

\bibitem[NeuralMagic(2022)]{deepsparse}
NeuralMagic.
\newblock {DeepSparse}, 2022.
\newblock URL \url{https://github.com/neuralmagic/deepsparse}.

\bibitem[Nilsback \& Zisserman(2006)Nilsback and
  Zisserman]{nilsback2006flowers}
Nilsback, M.-E. and Zisserman, A.
\newblock A visual vocabulary for flower classification.
\newblock In \emph{Conference on Computer Vision and Pattern Recognition
  (CVPR)}, 2006.

\bibitem[NVIDIA(2021)]{NVIDIA24}
NVIDIA.
\newblock {Accelerating Inference with Sparsity Using the NVIDIA Ampere
  Architecture and NVIDIA TensorRT}, 2021.
\newblock URL
  \url{https://developer.nvidia.com/blog/accelerating-inference-with-sparsity-using-ampere-and-tensorrt/}.

\bibitem[Park et~al.(2016)Park, Li, Wen, Tang, Li, Chen, and
  Dubey]{park2016faster}
Park, J., Li, S., Wen, W., Tang, P. T.~P., Li, H., Chen, Y., and Dubey, P.
\newblock Faster cnns with direct sparse convolutions and guided pruning.
\newblock \emph{arXiv preprint arXiv:1608.01409}, 2016.

\bibitem[Parkhi et~al.(2012)Parkhi, Vedaldi, Zisserman, and
  Jawahar]{parkhi2012apets}
Parkhi, O.~M., Vedaldi, A., Zisserman, A., and Jawahar, C.~V.
\newblock Cats and dogs.
\newblock In \emph{Conference on Computer Vision and Pattern Recognition
  (CVPR)}, 2012.

\bibitem[Paszke et~al.(2019{\natexlab{a}})Paszke, Gross, Massa, Lerer,
  Bradbury, Chanan, Killeen, Lin, Gimelshein, Antiga, Desmaison, Kopf, Yang,
  DeVito, Raison, Tejani, Chilamkurthy, Steiner, Fang, Bai, and
  Chintala]{pytorch}
Paszke, A., Gross, S., Massa, F., Lerer, A., Bradbury, J., Chanan, G., Killeen,
  T., Lin, Z., Gimelshein, N., Antiga, L., Desmaison, A., Kopf, A., Yang, E.,
  DeVito, Z., Raison, M., Tejani, A., Chilamkurthy, S., Steiner, B., Fang, L.,
  Bai, J., and Chintala, S.
\newblock {PyTorch}: An imperative style, high-performance deep learning
  library.
\newblock In \emph{Conference on Neural Information Processing Systems
  (NeurIPS)}. 2019{\natexlab{a}}.

\bibitem[Paszke et~al.(2019{\natexlab{b}})Paszke, Gross, Massa, Lerer,
  Bradbury, Chanan, Killeen, Lin, Gimelshein, Antiga,
  et~al.]{paszke2019pytorch}
Paszke, A., Gross, S., Massa, F., Lerer, A., Bradbury, J., Chanan, G., Killeen,
  T., Lin, Z., Gimelshein, N., Antiga, L., et~al.
\newblock Pytorch: An imperative style, high-performance deep learning library.
\newblock In \emph{Conference on Neural Information Processing Systems
  (NeurIPS)}, 2019{\natexlab{b}}.

\bibitem[Peste et~al.(2021)Peste, Iofinova, Vladu, and Alistarh]{peste2021ac}
Peste, A., Iofinova, E., Vladu, A., and Alistarh, D.
\newblock {AC/DC}: Alternating compressed/decompressed training of deep neural
  networks.
\newblock In \emph{Conference on Neural Information Processing Systems
  (NeurIPS)}, 2021.

\bibitem[Raihan \& Aamodt(2020)Raihan and Aamodt]{raihan2020sparse}
Raihan, M.~A. and Aamodt, T.
\newblock Sparse weight activation training.
\newblock \emph{Advances in Neural Information Processing Systems},
  33:\penalty0 15625--15638, 2020.

\bibitem[Rumelhart et~al.(1986)Rumelhart, Hinton, and
  Williams]{rumelhart1986learning}
Rumelhart, D.~E., Hinton, G.~E., and Williams, R.~J.
\newblock Learning representations by back-propagating errors.
\newblock \emph{nature}, 323\penalty0 (6088):\penalty0 533--536, 1986.

\bibitem[Russakovsky et~al.(2015)Russakovsky, Deng, Su, Krause, Satheesh, Ma,
  Huang, Karpathy, Khosla, Bernstein, et~al.]{imagenet}
Russakovsky, O., Deng, J., Su, H., Krause, J., Satheesh, S., Ma, S., Huang, Z.,
  Karpathy, A., Khosla, A., Bernstein, M., et~al.
\newblock Imagenet large scale visual recognition challenge.
\newblock \emph{International Journal of Computer Vision}, 115\penalty0
  (3):\penalty0 211--252, 2015.

\bibitem[Salman et~al.(2020)Salman, Ilyas, Engstrom, Kapoor, and
  Madry]{salman2020adversarially}
Salman, H., Ilyas, A., Engstrom, L., Kapoor, A., and Madry, A.
\newblock Do adversarially robust {ImageNet} models transfer better?
\newblock \emph{Conference on Neural Information Processing Systems (NeurIPS)},
  2020.

\bibitem[Sanh et~al.(2020)Sanh, Wolf, and Rush]{2020-sanh}
Sanh, V., Wolf, T., and Rush, A.~M.
\newblock Movement pruning: Adaptive sparsity by fine-tuning.
\newblock \emph{arXiv preprint arXiv:2005.07683}, 2020.

\bibitem[Schwarz et~al.(2021)Schwarz, Jayakumar, Pascanu, Latham, and
  Teh]{schwarz2021powerpropagation}
Schwarz, J., Jayakumar, S., Pascanu, R., Latham, P., and Teh, Y.
\newblock Powerpropagation: A sparsity inducing weight reparameterisation.
\newblock In \emph{Conference on Neural Information Processing Systems
  (NeurIPS)}, 2021.

\bibitem[Singh \& Alistarh(2020)Singh and Alistarh]{singh2020woodfisher}
Singh, S.~P. and Alistarh, D.
\newblock {WoodFisher}: Efficient second-order approximation for neural network
  compression.
\newblock In \emph{Conference on Neural Information Processing Systems
  (NeurIPS)}, 2020.

\bibitem[SparseZoo(2022)]{sparsezoo}
SparseZoo, N.
\newblock {DeepSparse}, 2022.
\newblock URL \url{https://github.com/neuralmagic/sparsezoo}.

\bibitem[Wang et~al.(2018)Wang, Singh, Michael, Hill, Levy, and
  Bowman]{wang2018glue}
Wang, A., Singh, A., Michael, J., Hill, F., Levy, O., and Bowman, S.~R.
\newblock Glue: A multi-task benchmark and analysis platform for natural
  language understanding.
\newblock \emph{arXiv preprint arXiv:1804.07461}, 2018.

\bibitem[Wang(2020)]{wang2020sparsert}
Wang, Z.
\newblock Sparsert: Accelerating unstructured sparsity on gpus for deep
  learning inference.
\newblock \emph{arXiv preprint arXiv:2008.11849}, 2020.

\bibitem[Wiedemann et~al.(2020)Wiedemann, Mehari, Kepp, and
  Samek]{wiedemann2020dithered}
Wiedemann, S., Mehari, T., Kepp, K., and Samek, W.
\newblock Dithered backprop: A sparse and quantized backpropagation algorithm
  for more efficient deep neural network training.
\newblock In \emph{Proceedings of the IEEE/CVF Conference on Computer Vision
  and Pattern Recognition Workshops}, pp.\  720--721, 2020.

\bibitem[Wolf et~al.(2019)Wolf, Debut, Sanh, Chaumond, Delangue, Moi, Cistac,
  Rault, Louf, Funtowicz, et~al.]{wolf2019huggingface}
Wolf, T., Debut, L., Sanh, V., Chaumond, J., Delangue, C., Moi, A., Cistac, P.,
  Rault, T., Louf, R., Funtowicz, M., et~al.
\newblock Huggingface's transformers: State-of-the-art natural language
  processing.
\newblock \emph{arXiv preprint arXiv:1910.03771}, 2019.

\bibitem[Xiao et~al.(2010)Xiao, Hays, Ehinger, Oliva, and
  Torralba]{xiao2010SUN}
Xiao, J., Hays, J., Ehinger, K., Oliva, A., and Torralba, A.
\newblock Sun database: Large-scale scene recognition from abbey to zoo.
\newblock \emph{Conference on Computer Vision and Pattern Recognition (CVPR)},
  2010.

\bibitem[Yang et~al.(2020)Yang, Ghasemazar, Ren, Golub, Lemieux, and
  Lis]{yang2020procrustes}
Yang, D., Ghasemazar, A., Ren, X., Golub, M., Lemieux, G., and Lis, M.
\newblock Procrustes: a dataflow and accelerator for sparse deep neural network
  training.
\newblock In \emph{2020 53rd Annual IEEE/ACM International Symposium on
  Microarchitecture (MICRO)}, pp.\  711--724. IEEE, 2020.

\bibitem[Zafrir et~al.(2021)Zafrir, Larey, Boudoukh, Shen, and
  Wasserblat]{zafrir2021prune}
Zafrir, O., Larey, A., Boudoukh, G., Shen, H., and Wasserblat, M.
\newblock Prune once for all: Sparse pre-trained language models.
\newblock \emph{arXiv preprint arXiv:2111.05754}, 2021.

\bibitem[Zhang et~al.(2020)Zhang, Yang, Ren, Su, and Sun]{zhang2020memorized}
Zhang, Z., Yang, P., Ren, X., Su, Q., and Sun, X.
\newblock Memorized sparse backpropagation.
\newblock \emph{Neurocomputing}, 415:\penalty0 397--407, 2020.

\bibitem[Zhu \& Gupta(2017)Zhu and Gupta]{zhu2017prune}
Zhu, M. and Gupta, S.
\newblock To prune, or not to prune: exploring the efficacy of pruning for
  model compression.
\newblock \emph{arXiv preprint arXiv:1710.01878}, 2017.

\end{thebibliography}
\bibliographystyle{icml2023}

%%%%%%%%%%%%%%%%%%%%%%%%%%%%%%%%%%%%%%%%%%%%%%%%%%%%%%%%%%%%%%%%%%%%%%%%%%%%%%%
%%%%%%%%%%%%%%%%%%%%%%%%%%%%%%%%%%%%%%%%%%%%%%%%%%%%%%%%%%%%%%%%%%%%%%%%%%%%%%%
% APPENDIX
%%%%%%%%%%%%%%%%%%%%%%%%%%%%%%%%%%%%%%%%%%%%%%%%%%%%%%%%%%%%%%%%%%%%%%%%%%%%%%%
%%%%%%%%%%%%%%%%%%%%%%%%%%%%%%%%%%%%%%%%%%%%%%%%%%%%%%%%%%%%%%%%%%%%%%%%%%%%%%%
\newpage
\appendix
%%%%%%%%%%%%%%%%%%%%%%%%%%%%%%%%%%%%%%%%%%%%%%%%%%%%%%%%%%%%%%%%%%%%%%%%%%%%%%%
%%%%%%%%%%%%%%%%%%%%%%%%%%%%%%%%%%%%%%%%%%%%%%%%%%%%%%%%%%%%%%%%%%%%%%%%%%%%%%%

\section{Additional Detailed Results}

In this section, we describe the twelve datasets we use to train image classification models in sections \ref{subsection:transfer} and \ref{subsection:scratch}, as well as present the complete per-dataset accuracy results for transfer and from-scratch training on these datasets.

\begin{table}[h!]
\centering
\scalebox{0.48}{
\begin{tabular}{@{}ccccc@{}}
\toprule
Dataset & Number of Classes &  Train/Test Examples & Accuracy Metric\\
\midrule
SUN397\cite{xiao2010SUN} & 397 & 19 850 / 19 850 & Top-1 \\
FGVC Aircraft\cite{maji13fgvc-aircraft} & 100 & 6 667 / 3 333 & Mean Per-Class \\
Birdsnap\cite{berg2014birds} & 500 & 32 677 / 8 171 & Top-1 \\
Caltech-101\cite{li2004caltech101} & 101 & 3 030 / 5 647 & Mean Per-Class \\
Caltech-256\cite{holub2006caltech256} & 257 & 15 420 / 15 187 & Mean Per-Class \\
Stanford Cars\cite{krause2013cars} & 196 & 8 144 / 8 041 & Top-1 \\
CIFAR-10\cite{cifar100} & 10 & 50 000 / 10 000 & Top-1 \\
CIFAR-100\cite{cifar100}  & 100 & 50 000 / 10 000 & Top-1 \\
Describable Textures (DTD)\cite{cimpoi2014dtd} & 47 & 3 760 / 1 880 & Top-1 \\
Oxford 102 Flowers\cite{nilsback2006flowers} & 102 & 2 040 / 6 149 & Mean Per-Class \\
Food-101\cite{bossard2014food101} & 101 & 75 750 / 25 250 & Top-1 \\
Oxford-IIIT Pets\cite{parkhi2012apets} & 37 & 3 680 / 3 669 & Mean Per-Class \\
\bottomrule
\end{tabular}

}
\caption{Target tasks for from-scratch and transfer learning.}
\label{table:datasets}
\end{table}

\begin{table}[h]
\centering
\scalebox{0.65}{%\
\begin{tabular}{ccccc}
\toprule
{Dataset} & {Dense} & {Uniform 90\%} & {Uniform 97\%} & {Global 95\%}  \\
\midrule
Aircraft & 83.6 $ \pm $ 0.4   & 81.4 $\pm$ 0.3 & 79.0 $\pm$ 0.0 & 81.2 $ \pm $ 0.4 \\
Birds & 72.4 $ \pm $ 0.3 & 68.7 $\pm$ 0.1 & 67.8 $\pm$ 0.0 & 66.9 $ \pm $ 0.1  \\
CIFAR-10 & 97.4 $ \pm $ 0.0 & 97.0 $\pm$ 0.0 &  96.7 $\pm$ 0.3 & 96.2 $ \pm $ 0.1 \\
CIFAR-100 & 85.6 $ \pm $ 0.2 & 84.5 $\pm$ 0.1 & 84.0 $\pm$ 0.1 & 82.9 $ \pm $ 0.1 \\
Caltech-101 & 93.5 $ \pm $ 0.1 & 92.5 $\pm$ 0.1 & 92.1 $\pm$ 0.3 & 91.9 $ \pm $ 0.2  \\
Caltech-256 & 86.1 $ \pm $ 0.1 & 85.1 $\pm$ 0.0 & 83.6 $\pm$ 0.0 & 83.1 $ \pm $ 0.0 \\
Cars & 90.3 $ \pm $ 0.2 & 88.2 $\pm$ 0.2 & 87.0 $\pm$ 0.1 & 87.6 $ \pm $ 0.1 \\
DTD & 76.2 $ \pm $ 0.3 & 75.1 $\pm$ 0.0 & 74.8 $\pm$ 0.2 & 74.1 $ \pm $ 0.4  \\
Flowers & 95.0 $ \pm $ 0.1 & 95.0 $\pm$ 0.0  & 95.3 $\pm$ 0.4 & 94.1 $ \pm $ 0.3 \\
Food-101 & 87.3 $ \pm $ 0.1 & 86.5 $\pm$ 0.1 & 85.7 $\pm$ 0.0 & 85.5 $ \pm $ 0.0 \\
Pets & 93.4 $ \pm $ 0.1 & 92.3 $\pm$ 0.1 & 90.1 $\pm$ 0.0 & 91.0 $ \pm $ 0.1\\
SUN397 & 64.8 $ \pm $ 0.0 & 63.4 $\pm$ 0.0 & 62.4 $\pm$ 0.1 & 61.4 $ \pm $ 0.2\\

\bottomrule
\end{tabular}

}
\caption{Transfer accuracy for sparse ResNet50 models pretrained on ImageNet1K.}
\label{table:rn50_full_all}
\end{table}

\begin{table}[h]
\centering
\scalebox{0.65}{%\
\begin{tabular}{ccccc}
\toprule
{Dataset} & {Dense} & {Uniform 90 \%} & {Uniform 95\%} & {Global 95\%}  \\
\midrule
Aircraft & 58.1 $ \pm $ 0.3 & 56.1 $\pm$ 0.4 & 56.0 $\pm$ 0.1 & 57.6 $ \pm $ 0.3 \\
Birds & 51.5 $ \pm $ 0.7 & 50.3 $\pm$ 0.3 & 49.2 $\pm$ 0.5 & 50.1 $ \pm $ 0.3  \\
CIFAR-10 & 94.3 $ \pm $ 0.3 & 93.7 $\pm$ 0.1 &  93.1 $\pm$ 0.2 & 93.5 $ \pm $ 0.1 \\
CIFAR-100 & 74.6 $ \pm $ 0.0 & 73.4 $\pm$ 0.3 & 72.8 $\pm$ 0.3 & 72.8 $ \pm $ 0.6 \\
Caltech-101 & 46.4 $ \pm $ 0.9 & 45.7  & 45.0 & 45.6 $ \pm $ 0.8  \\
Caltech-256 & 47.1 $ \pm $ 0.8 & 46.2 $\pm$ 0.8 & 45.7 $\pm$ 0.7 & 46.4 $ \pm $ 0.8 \\
Cars & 67.1 $ \pm $ 0.1 & 63.8 $\pm$ 0.5 & 62.7 $\pm$ 0.8 & 64.5 $ \pm $ 0.5 \\
DTD & 37.8 $ \pm $ 1.3 & 36.3 $\pm$ 1.5 & 37.0 $\pm$ 1.0 & 37.7 $ \pm $ 1.1  \\
Flowers & 59.3 $ \pm $ 0.1 & 59.4 $\pm$ 0.5  & 58.5 $\pm$ 0.8 & 58.2 $ \pm $ 0.4 \\
Food-101 & 78.3 $ \pm $ 1.7 & 77.1 $\pm$ 0.0 & 76.0 $\pm$ 0.2 & 77.4 $ \pm $ 0.3 \\
Pets & 59.2 $ \pm $ 0.2 & 58.7 $\pm$ 0.3 & 57.3 $\pm$ 0.6 & 59.1 $ \pm $ 0.6\\
SUN397 & 40.5 $ \pm $ 3.2 & 40.0 $\pm$ 0.4 & 39.5 $\pm$ 0.1 & 39.5 $ \pm $ 0.1\\

\bottomrule
\end{tabular}

}
\caption{From-scratch training accuracy for sparse ResNet18 Models trained on standard training datasets.}
\label{table:from_scratch_acc}
\end{table}

\end{document}